\documentclass[preprint,review,12pt]{elsarticle}

\usepackage{amsmath,amsfonts,bm}

\def\eqref#1{equation~\ref{#1}}

\def\1{\bm{1}}

\DeclareMathAlphabet{\mathsfit}{\encodingdefault}{\sfdefault}{m}{sl}
\SetMathAlphabet{\mathsfit}{bold}{\encodingdefault}{\sfdefault}{bx}{n}

\usepackage{hyperref}
\usepackage{url}
\usepackage{siunitx}
\usepackage{tabularx}
\usepackage{amssymb}
\usepackage{amsmath}
\usepackage{float}
\usepackage[numbers]{natbib}
\usepackage{algorithm}
\usepackage{xcolor}
\usepackage{graphicx}
\usepackage{subcaption}
\usepackage{booktabs}
\usepackage{tikz}
\usepackage{caption}
\usepackage{algorithmicx}
\usepackage{algpseudocode} 
\usepackage[a4paper, total={6.5in, 8in}]{geometry}

\begin{document}

\begin{frontmatter}

\title{Transformer-Based Reinforcement Learning for Autonomous Orbital Collision Avoidance in Partially Observable Environments.} 

\author[1]{Thomas Georges}

\author[1]{Adam Abdin\corref{cor1}} 

\cortext[cor1]{Corresponding author: 
  E-Mail: adam.abdin@centralesupelec.fr (Adam Abdin)}
\affiliation[1]{organization={Université Paris-Saclay, CentraleSupélec Engineering School, Department of Industrial Engineering},
                city={Gif-sur-Yvette},
                postcode={91190},
                country={France}}

\begin{abstract}

We introduce a Transformer-based Reinforcement Learning framework for autonomous orbital collision avoidance that explicitly models the effects of partial observability and imperfect monitoring in  space operations. The framework combines a configurable encounter simulator, a distance-dependent observation model, and a sequential state estimator to represent uncertainty in relative motion. A central contribution of this work is the use of transformer-based Partially Observable Markov Decision Process (POMDP) architecture, which leverage long-range temporal attention to interpret noisy and intermittent observations more effectively than traditional architectures. This integration provides a foundation for training collision avoidance agents that can operate more reliably under imperfect monitoring environments.

\end{abstract}


\end{frontmatter}



\section{Introduction}
\label{sec:intro}

The proliferation of satellites in Low Earth Orbit (LEO) is transforming collision avoidance from an occasional operational concern into a persistent operational challenge. With over 8,000 active satellites currently in orbit and projected growth to exceed 60,000 by 2030 due to mega-constellations, the frequency of conjunction events requiring assessment and management is projected to grow substantially \cite{nasaCAHandbook2023,li2022review}. Each potential collision demands rapid decision-making that must balance collision risk against fuel consumption and mission continuity.

Current collision avoidance operations rely primarily on  ground-based decision-making pipelines. Space surveillance networks track objects, propagate trajectories, compute collision probabilities, and generate maneuver recommendations that are uploaded to spacecraft \cite{Pavanello2025CAMmary}. This approach has proven effective and has prevented numerous collisions. Nontheless, it faces scalability limitations as the number of tracked objects and managed satellites grows. The computational burden of screening conjunction pairs increases rapidly, while communication delays between ground stations and spacecraft can exceed the time available for effective maneuvering. Scaling these operations to future traffic levels will require adapted tools and intelligence-assisted decision-making frameworks.

A central challenge in decision-making for collision avoidance is imperfect monitoring. Space surveillance networks provide tracking data that is inherently imperfect: coverage is intermittent, measurement noise varies with sensor geometry and atmospheric conditions, and state estimates degrade progressively as time elapses since the last observation. On-board satellite sensors face their own limitations in range, accuracy, and operational availability. Maneuver decisions must therefore be made based on information that is partial and of uncertain quality \cite{Langsdale2025,Wong2016}. 

When reliable state estimates are available, optimization-based approaches to maneuver planning have proven effective. Convex programming and nonlinear optimization techniques can generate fuel-efficient maneuvers that achieve a desired miss distance or reduce collision probability below acceptable thresholds \cite{Malyuta2019,Pavanello2024-poly}. These methods perform well when the conjunction geometry is known with confidence and the planning problem can be treated as a single optimization instance. Collision avoidance is, however, fundamentally sequential in nature. A maneuver executed at one time impacts the spacecraft trajectory, which in turn alters the geometry of subsequent potential conjunctions. Information arrives incrementally as new tracking observations become available. The decision of whether to maneuver immediately or wait for better information carries consequences that propagate forward in time. This temporal structure is not fully captured by single-shot static optimization.

Sequential and dynamic decision-making frameworks address this limitation effectively. Model Predictive Control (MPC) solves finite-horizon optimization problems repeatedly as new information arrives, providing a natural mechanism for incorporating updated state estimates \cite{Sanchez2025-ccmpc}. Recent extensions handle multi-encounter scenarios and actuator constraints while maintaining computational tractability through convex relaxations \cite{Pavanello2024-multienc}. On the other hand, Reinforcement learning (RL) offers an alternative paradigm, in which policies mapping states to actions are learned through interaction with simulated environments without requiring explicit trajectory optimization at decision time.

These sequential methods share a common assumption that the state is known or that uncertainty can be adequately captured through conservative bounds. MPC requires a state estimate to initialize each planning horizon. Standard RL formulations assume the agent observes the true state or a fixed noisy transformation of it. When state uncertainty is substantial, these methods tend toward excessive conservatism, executing larger maneuvers than necessary. When uncertainty is underestimated, they risk failing to avoid collisions altogether \cite{Ryu2024,Elango2025}.

The Partially Observable Markov Decision Process (POMDP) framework provides a principled foundation for decision-making when states cannot be directly observed \cite{Kaelbling1998}. A POMDP policy maps observation histories to actions, implicitly or explicitly maintaining a belief distribution over possible states. This framework has found application in aerospace domains where sensor limitations preclude full observability. \citet{Bai2011} demonstrated that belief-space planning substantially outperforms state-based approaches for unmanned aircraft collision avoidance under sensor uncertainty. The critical insight is that optimal actions depend on the shape of the entire uncertainty distribution, not only on the most likely state estimate.

However, due to tractability issues and computational complexity, solving POMDPs with continuous state and action spaces requires approximation. Deep reinforcement learning provides one approach: neural networks can learn policies that implicitly aggregate observation histories into representations sufficient for action selection \cite{Hausknecht2015DRQN,Igl2018}. The network architecture determines how information from past observations is retained and combined. Recurrent neural networks, particularly Long Short-Term Memory (LSTM) networks, maintain a hidden state that is updated with each new observation \cite{Kapturowski2019R2D2}. This hidden state compresses the observation history into a fixed-dimensional vector from which actions are selected. While effective in many domains, recurrent architectures can struggle with long sequences and exhibit sensitivity to the timing and spacing of observations.

Alternative architectures have been explored for memory-augmented RL. Memory networks with explicit read-write operations allow selective storage and retrieval of past information \cite{Parisotto2020}. State-space models provide structured approaches to temporal modeling with computational advantages for long sequences \cite{Gu2022}. Each architecture embodies different assumptions about what information from the past is relevant and how it should be accessed.

More recently, the Transformer architecture offers a distinct approach to temporal aggregation \cite{vaswani2017attention}. Rather than compressing history into a recurrent state, transformers maintain representations of individual past observations and use attention mechanisms to selectively weight their influence on current decisions. This selective attention is particularly relevant when observation quality varies over time and when useful information may be separated by intervals of uninformative or missing data. Transformer-XL extends this baseline architecture to handle sequences longer than the training context through segment-level recurrence \cite{Dai2019TXL}. Recent work has demonstrated the effectiveness of transformer-based policies for RL in partially observable environments \cite{chen2024transdreamerreinforcementlearningtransformer}, however, these studies typically assume stationary observation noise. In orbital collision avoidance, observation quality depends on the state itself in which sensing degrades with distance and improves as conjunction approaches, creating a non-stationary information structure that the policy must learn to exploit.

To address these gaps, this paper develops a RL framework for spacecraft collision avoidance that treats partial observability as a central modeling concern. The observation model reflects operational considerations such as variation in information quality, and observations becoming noisier and more prone to dropout as separation distance increases. We employ an Unscented Kalman Filter (UKF) for sequential state estimation and use its covariance output to define a reward based on the Mahalanobis distance, which weights miss distance by estimation uncertainty. The policy architecture uses Transformer-XL to aggregate noisy and intermittent observations over time, producing maneuver decisions that respond to the actual information content available at each decision epoch.

The contributions of this work are as follows:

\begin{enumerate}
\item A POMDP formulation for collision avoidance that incorporates partial observability as an explicit function of geometry, with smooth interpolation across sensing regimes maintaining differentiability for policy learning.

\item A reward formulation combining collision probability estimation with a Mahalanobis distance-based surrogate, providing stable gradients for policy optimization while remaining grounded in operationally meaningful risk metrics.

\item The application of Transformer-XL architecture to spacecraft collision avoidance, demonstrating consistent improvement in fuel efficiency over memoryless baselines while maintaining collision avoidance performance across tested observability conditions.

\item Systematic evaluation across five observability regimes, from near-perfect sensing to severe degradation, characterizing how learned policies adapt their behavior to information quality.
\end{enumerate}

The remainder of this paper is organized as follows. Section \ref{sec:problem} formulates the collision avoidance problem as a Partially Observable Markov Decision Process (POMDP). Section \ref{sec:methodology} presents the methodology, including orbital dynamics, state estimation, reward design, and policy architecture. Section \ref{sec:results} presents the results and the evaluation of performance across observability regimes. It also compares transformer-based policies against feedforward baselines. Section \ref{sec:conclusion} concludes with implications for operational deployment and directions for future work.


\section{Problem Formulation}
\label{sec:problem}

\subsection{The Collision Avoidance Decision Problem}

We consider the problem of collision avoidance between a maneuverable spacecraft and debris objects in Earth orbit. At discrete decision epochs $t \in \{0, \Delta t, 2\Delta t, \ldots, T\}$, the spacecraft must determine whether to execute an orbital maneuver to reduce collision risk. The objective is to minimize the probability of collision while limiting fuel consumption and maintaining the spacecraft's intended trajectory. Each maneuver incurs a fuel cost and perturbs the orbit, while failing to maneuver when necessary risks collision. The problem is further complicated by the requirement to plan over a finite horizon, as maneuvers affect not only immediate collision risk but also future encounter geometries.

\subsection{State Space and Dynamics}

The state at time $t$ consists of the relative position and velocity between the spacecraft and debris in the Local Vertical Local Horizontal (LVLH) reference frame centered on the spacecraft. The LVLH frame is defined with basis vectors $\{\hat{\mathbf{R}}, \hat{\mathbf{T}}, \hat{\mathbf{W}}\}$, where $\hat{\mathbf{R}}$ points radially outward from Earth, $\hat{\mathbf{T}}$ aligns with the velocity vector in the orbital plane, and $\hat{\mathbf{W}}$ completes the right-handed coordinate system, pointing normal to the orbital plane.

The relative state vector is defined as:
\begin{equation}
\mathbf{x}_t = \begin{bmatrix} \delta\mathbf{r}_t \\ \delta\mathbf{v}_t \end{bmatrix} \in \mathbb{R}^6,
\label{eq:state}
\end{equation}
where $\delta\mathbf{r}_t \in \mathbb{R}^3$ represents the relative position and $\delta\mathbf{v}_t \in \mathbb{R}^3$ represents the relative velocity at decision epoch $t$.

The state evolution follows nonlinear two-body Keplerian dynamics. Let $\mathbf{s}_t^{(S)} = [\mathbf{r}_t^{(S)}; \mathbf{v}_t^{(S)}]$ and $\mathbf{s}_t^{(D)} = [\mathbf{r}_t^{(D)}; \mathbf{v}_t^{(D)}]$ denote the inertial states of the spacecraft and debris respectively. The dynamics are governed by:
\begin{align}
\mathbf{s}_{t+1}^{(S)} &= \Phi_{\text{Kep}}\left(\mathbf{s}_t^{(S)}, \Delta\mathbf{v}_t, \Delta t\right) + \boldsymbol{\eta}_t^{(S)}, \label{eq:dynamics_spacecraft}\\
\mathbf{s}_{t+1}^{(D)} &= \Phi_{\text{Kep}}\left(\mathbf{s}_t^{(D)}, \mathbf{0}, \Delta t\right) + \boldsymbol{\eta}_t^{(D)}, \label{eq:dynamics_debris}
\end{align}
where $\Phi_{\text{Kep}}(\cdot)$ represents the two-body orbital propagation operator, $\Delta\mathbf{v}_t \in \mathbb{R}^3$ is the velocity change imparted by the maneuver at time $t$, and $\boldsymbol{\eta}_t^{(\cdot)}$ captures process noise including unmodeled perturbations. The relative state evolution can be derived from these inertial dynamics through coordinate transformation to the LVLH frame.

The action space consists of impulsive velocity changes $\mathbf{a}_t = \Delta\mathbf{v}_t \in \mathcal{A} \subset \mathbb{R}^3$, where $\mathcal{A}$ is bounded by thrust capabilities: $\|\Delta\mathbf{v}_t\| \leq \Delta v_{\max}$. These maneuvers instantaneously change the spacecraft velocity while the debris continues on its uncontrolled trajectory.

\subsection{Partial Observability}

The spacecraft has imperfect information about the true state $\mathbf{x}_t$. Rather than observing the state directly, the spacecraft receives noisy observations:
\begin{equation}
\mathbf{o}_t = h_t(\mathbf{x}_t) + \mathbf{v}_t, \quad \mathbf{v}_t \sim \mathcal{N}(\mathbf{0}, \mathbf{R}_t),
\label{eq:observation}
\end{equation}
where $h_t: \mathbb{R}^6 \rightarrow \mathbb{R}^m$ is the observation function, $\mathbf{v}_t$ represents measurement noise, and $\mathbf{R}_t$ is the observation noise covariance matrix. The observation function $h_t$ may vary over time due to changing geometric configurations and sensor availability.

A critical aspect of the problem is that observation quality degrades with the distance between spacecraft and debris. Both the measurement noise covariance and the probability of observation dropout increase with separation distance $d_t = \|\delta\mathbf{r}_t\|$. This reflects operational conditions where tracking accuracy depends on geometry and range. Observations may also be intermittent, with certain state components unavailable at various decision epochs.

\subsection{POMDP Formulation}

We formulate collision avoidance as a Partially Observable Markov Decision Process (POMDP):
\begin{equation}
\mathcal{M} = (\mathcal{S}, \mathcal{A}, \mathcal{T}, \mathcal{R}, \Omega, \mathcal{O}, \gamma),
\end{equation}
where:

\begin{itemize}
\item $\mathcal{S} = \mathbb{R}^6$ is the continuous state space representing all possible relative positions and velocities.
\item $\mathcal{A} \subset \mathbb{R}^3$ is the continuous action space of feasible velocity changes bounded by thrust constraints.
\item $\mathcal{T}: \mathcal{S} \times \mathcal{A} \times \mathcal{S} \rightarrow [0,1]$ is the state transition probability kernel induced by the stochastic dynamics in Equations~\ref{eq:dynamics_spacecraft}--\ref{eq:dynamics_debris}.
\item $\mathcal{R}: \mathcal{S} \times \mathcal{A} \rightarrow \mathbb{R}$ is the reward function encoding the trade-off between collision risk, fuel consumption, and trajectory deviation.
\item $\Omega$ is the observation space containing all possible measurements.
\item $\mathcal{O}: \mathcal{S} \times \Omega \rightarrow [0,1]$ is the observation probability function determined by Equation~\ref{eq:observation}.
\item $\gamma \in (0,1]$ is the discount factor reflecting the planning horizon.
\end{itemize}

The solution to this POMDP is a policy $\pi: \mathcal{H} \rightarrow \mathcal{A}$ that maps the history of observations $\mathcal{H}_t = \{\mathbf{o}_1, \ldots, \mathbf{o}_t\}$ to actions. Since the state is not directly observable, the spacecraft must maintain a belief state $b_t(\mathbf{x})$ representing the probability distribution over possible states given the observation history.

\subsection{Objectives and Constraints}

A collision occurs when the relative distance falls below the combined hard-body radius: $\|\delta\mathbf{r}_t\| < r_{\text{combined}}$, where $r_{\text{combined}} = r_{\text{spacecraft}} + r_{\text{debris}}$. The primary objective is to maintain the probability of collision $P_c(t)$ below an acceptable threshold $P_c^{\text{max}}$ throughout the encounter.

The decision problem involves three competing objectives:
\begin{enumerate}
\item Safety: Minimize the collision probability $P_c(t)$ over the planning horizon.
\item Fuel efficiency: Minimize the total velocity change $\sum_{t=0}^{T} \|\Delta\mathbf{v}_t\|$ to preserve propellant for future operations.
\item Trajectory maintenance: Minimize deviations from the nominal orbital elements to maintain mission objectives.
\end{enumerate}

These objectives are conflicting since conservative maneuvering reduces collision risk but increases fuel consumption, while delayed maneuvering preserves fuel but may inadequately address collision risk given state uncertainty. The relative importance of these objectives is encoded in the reward function, which must also account for uncertainty in state estimation when evaluating collision risk.


\section{Methodology}
\label{sec:methodology}
\subsection{Orbital Dynamics and Trajectory Modelling}

The on-orbit motion of the spacecraft and debris follows the two-body Keplerian dynamics described in Section~\ref{sec:problem}. The relative state $\mathbf{x}_t$ evolves according to Equations~\ref{eq:dynamics_spacecraft}--\ref{eq:dynamics_debris}, with all relative quantities expressed in the LVLH frame defined in Section~\ref{sec:problem}.

To generate collision-prone yet physically consistent training episodes, the framework employs an inverse-encounter procedure that maps a user-specified encounter descriptor to Keplerian initial conditions. An encounter descriptor comprises a target miss distance $d_{\mathrm{miss}}$, a target relative speed $v_{\mathrm{rel}}$ at closest approach, and an approach direction in LVLH parameterised by angles $(\psi,\phi)$. An optional in-plane yaw $\psi_{\mathrm{yaw}}$ fixes the phase on the encounter ellipse.

The generator first samples a feasible orbit for the spacecraft, specifying semi-major axis, eccentricity, inclination, argument of perigee, right ascension of the ascending node, and mean anomaly. It then selects a debris orbit and phases it so that the induced relative motion satisfies
\begin{equation}
t^\star = \arg\min_{t\in[0,T]} \| \delta \mathbf{r}_t \|, \qquad 
\| \delta \mathbf{r}_{t^\star} \| \approx d_{\mathrm{miss}}, \qquad 
\| \delta \mathbf{v}_{t^\star} \| \approx v_{\mathrm{rel}},
\label{eq:encounter-constraints}
\end{equation}
with approach direction aligned to $(\psi,\phi)$ and in-plane yaw $\psi_{\mathrm{yaw}}$. Start and end separations are constrained so that $\|\delta \mathbf{r}_0\|\in[d_{\min}^{\mathrm{start}}, d_{\max}^{\mathrm{start}}]$ and $\|\delta \mathbf{r}_T\|\in[d_{\min}^{\mathrm{end}}, d_{\max}^{\mathrm{end}}]$, and a minimum time buffer is reserved around $t^\star$ to ensure sufficient maneuvering opportunity. Initial conditions $(\mathbf{s}^{(S)}_0, \mathbf{s}^{(D)}_0)$ are accepted if tolerances on the encounter parameters are met; otherwise, phases and debris elements are resampled.

\begin{figure}[ht]
    \includegraphics[width=\textwidth]{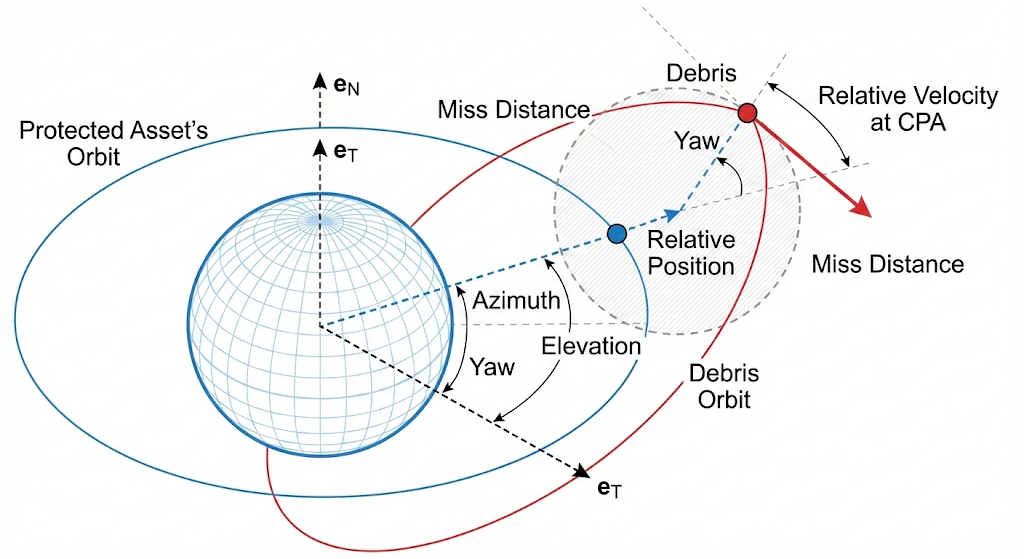}
    \caption{Schematic of the inverse-encounter trajectory generation procedure.}
    \label{fig:diagram_trajectory}
\end{figure}

Figure~\ref{fig:diagram_trajectory} illustrates the trajectory generation procedure. All encounter parameters are configurable, enabling systematic sweeps over relative speed, geometry, miss distance, and time-to-closest-approach. Using fixed random seeds reproduces identical Keplerian ephemerides, mask realisations, and encounter geometries, facilitating controlled comparisons across architectures and training configurations.

\subsection{Distance-Dependent Partial Observability via Lagrange Interpolation}
\label{subsec:ddpo}

Sensor quality degrades with range. To capture this in a controllable and differentiable manner, the observation noise scale $\sigma(d)$ and a masking probability $M(d)$ are modelled as explicit functions of the instantaneous separation $d = \|\delta\mathbf{r}_t\|$ between spacecraft and debris.

Four anchor distances specify the desired sensing-quality profile across regimes.
In the implementation, both the noise scale $\sigma(d)$ and masking probability $M(d)$
are obtained by exact cubic polynomial interpolation in $\log_{10}(d/\mathrm{km})$.
To enforce valid ranges, we interpolate $\log \sigma$ (and exponentiate) and interpolate
$\mathrm{logit}(M)$ (and apply a sigmoid), yielding $\sigma(d)>0$ and $M(d)\in(0,1)$.

Given the true state $\mathbf{x}_t$, we first apply distance-dependent multiplicative Gaussian noise and then (with probability $M(d_t)$) mask a configured subset of coordinates by carrying forward their previous observed values (hold-last dropout):
\begin{equation}
\label{eq:PO}
\begin{aligned}
\tilde{\mathbf{o}}_t &= \mathbf{x}_t \odot \left(1 + \boldsymbol{\epsilon}_t\right),
\qquad \boldsymbol{\epsilon}_t \sim \mathcal{N}\!\left(\mathbf{0},\,\sigma(d_t)^2 I\right),\\
\mathbf{o}_{t,k} &= 
\begin{cases}
\tilde{\mathbf{o}}_{t,k}, & \text{if coordinate $k$ is not masked at $t$},\\
\mathbf{o}_{t-1,k}, & \text{if coordinate $k$ is masked at $t$}.
\end{cases}
\end{aligned}
\end{equation}

This formulation provides two mechanisms for information degradation: multiplicative
noise with scale $\sigma(d_t)$ applied to the state components, and hold-last dropout,
where a configured subset of components is replaced by its previous observed value
with probability $M(d_t)$. Both effects intensify as separation grows, reflecting
the operational reality that tracking and sensing become less reliable at longer
ranges.

\subsection{Policy Optimization}

The collision avoidance problem is formulated as a sequential decision process in which the decision agent selects actions based on observations to maximize cumulative reward. In the fully observable case, this corresponds to a Markov Decision Process (MDP) defined by the tuple $(\mathcal{S}, \mathcal{A}, p, r, \gamma)$, where $p(\mathbf{s}_{t+1} \mid \mathbf{s}_t, \mathbf{a}_t)$ is the transition kernel, $r(\mathbf{s}_t, \mathbf{a}_t)$ is the reward function, and $\gamma \in (0,1)$ is the discount factor. The objective is to find a policy $\pi_\theta(\mathbf{a}_t \mid \mathbf{x}_t)$ that maximizes the expected discounted return $J(\theta) = \mathbb{E}_{\pi_\theta}\left[\sum_{t=0}^{T-1} \gamma^t r_t\right]$ \cite{SuttonBarto2018}.

Policy gradient methods optimize $J(\theta)$ directly by estimating the gradient with respect to policy parameters. The policy gradient theorem expresses this gradient as:
\begin{equation}
\nabla_\theta J(\theta) = \mathbb{E}_{\pi_\theta}\left[\nabla_\theta \log \pi_\theta(\mathbf{a}_t \mid \mathbf{x}_t) \, \widehat{A}_t\right],
\end{equation}
where $\widehat{A}_t$ estimates the advantage function $A^\pi(\mathbf{s}_t, \mathbf{a}_t) = Q^\pi(\mathbf{s}_t, \mathbf{a}_t) - V^\pi(\mathbf{s}_t)$ \cite{Williams1992,Sutton1999}. The advantage quantifies the relative merit of an action compared to the average performance from that state. In this work, advantages are estimated using Generalized Advantage Estimation (GAE):
\begin{equation}
\widehat{A}_t = \sum_{l=0}^{\infty} (\gamma \lambda)^l \delta_{t+l}, \qquad \delta_t = r_t + \gamma V_\phi(\mathbf{s}_{t+1}) - V_\phi(\mathbf{s}_t),
\end{equation}
where $\lambda \in [0,1]$ controls the bias-variance trade-off and $V_\phi$ is a learned value function \cite{Schulman2016GAE}.

Proximal Policy Optimization (PPO) stabilizes policy gradient updates by constraining the magnitude of policy changes at each iteration \cite{schulman2017}. The algorithm optimizes a clipped surrogate objective:
\begin{equation}
L^{\mathrm{CLIP}}(\theta) = \mathbb{E}\left[\min\left(\rho_t(\theta) \, \widehat{A}_t, \; \mathrm{clip}(\rho_t(\theta), 1-\epsilon, 1+\epsilon) \, \widehat{A}_t\right)\right],
\end{equation}
where $\rho_t(\theta) = \pi_\theta(\mathbf{a}_t \mid \mathbf{x}_t) / \pi_{\theta_{\text{old}}}(\mathbf{a}_t \mid \mathbf{x}_t)$ is the probability ratio between the current and previous policies, and $\epsilon$ is a clipping parameter that limits the effective update size. The policy is parameterized as a Gaussian distribution whose mean specifies the commanded $\Delta\mathbf{v}_t$, enabling continuous control with end-to-end differentiability from observations to actions.

\subsection{State Estimation and Collision Risk}
\label{subsec:collision_risk}

This section presents a novel collision risk surrogate that couples sequential state estimation with Mahalanobis distance for use in the proposed RL framework.

Conjunction analysis requires estimating the probability that two objects occupy the same region during a close approach. Let $\mathbf{r}_t = \mathbf{r}^{(D)}_t - \mathbf{r}^{(S)}_t$ denote the relative position, with mean $\boldsymbol{\mu}_t = \mathbb{E}[\mathbf{r}_t]$ and covariance $\Sigma_{r,t} = \mathrm{Cov}(\mathbf{r}_t)$. For short-duration encounters, the standard treatment projects these statistics onto the encounter plane, spanned by the along-track and cross-track directions and normal to the relative velocity at the time of closest approach $t_c$. Let $\boldsymbol{\mu}_{TW}$ and $\Sigma_{TW}$ denote the projected mean and covariance. The collision probability reduces to the integral of a bivariate normal over a disk of radius $r_A$ equal to the combined hard-body radius:
\begin{equation}
  P_c = \iint_{x^2+y^2 \le r_A^2} \varphi_2\left([x,y]^\top; \boldsymbol{\mu}_{TW}, \Sigma_{TW}\right) \mathrm{d}x\,\mathrm{d}y,
  \label{eq:pc_integral}
\end{equation}
where $\varphi_2(\cdot; \boldsymbol{\mu}, \Sigma)$ is the bivariate Gaussian density. Various methods exist for evaluating this integral, differing primarily in computational approach and numerical stability across different covariance geometries \cite{Patera2001,alfano2005,chan2008,li2022review}.

The quality of the covariance estimate $\Sigma_{r,t}$ is critical. Assuming independent estimation errors for the spacecraft and debris, the relative-position covariance is $\Sigma_{r,t} \approx \Sigma^{(D)}_{r,t} + \Sigma^{(S)}_{r,t}$. The encounter-plane covariance follows from projection at $t_c$: $\Sigma_{TW} = R_{TW} \Sigma_{r,t_c} R_{TW}^\top$. Since $P_c$ is a nonlinear function of $\Sigma_{TW}$, small errors in variance estimation can shift a conjunction across operational thresholds, causing either false alarms or missed detections.

To obtain the relative-position covariance used in the risk computation, we employ an
Unscented Kalman Filter (UKF) \cite{julier2004,wan2000} on the 6D relative Cartesian state
$\mathbf{x}_t = [\delta \mathbf{r}_t;\, \delta \mathbf{v}_t]$. In our implementation the
process model is a constant-velocity kinematic update over timestep $\Delta t$, and the
measurement model observes relative position. We use the resulting $3\times 3$ covariance
of $\delta\mathbf{r}_t$ as $\Sigma_{r,t}$ for collision-probability evaluation and for the
Mahalanobis-based surrogate.

For evaluating Equation~\eqref{eq:pc_integral}, we use the Chen--Bai short-encounter
approximation \cite{chen2017}. In implementation we form the encounter-plane mean and
covariance by projection and evaluate the resulting analytic expression in a numerically
stable form (including an \texttt{expm1}-based computation of $1-e^{-x}$ for small $x$),
avoiding the numerical sensitivity of naive quadrature.

While $P_c$ is the operationally meaningful quantity, using it directly as a reinforcement learning reward presents difficulties. The function $P_c(\boldsymbol{\mu}_{TW}, \Sigma_{TW})$ is extremely flat when objects are far apart and sharply curved near the hard-body boundary, leading to high-variance gradient estimates. To obtain well-behaved gradients, we introduce a surrogate based on the squared Mahalanobis distance in the encounter plane:
\begin{equation}
  d_M^2 = \boldsymbol{\mu}_{TW}^\top \Sigma_{TW}^{-1} \boldsymbol{\mu}_{TW}, \qquad
  \tilde{r}_{\text{safe}} = \exp\left(-\tfrac{1}{2} d_M^2\right).
  \label{eq:maha}
\end{equation}
This mapping is smooth in both the mean and covariance, strictly monotone along covariance ellipses, and preserves the level sets of Equation~\eqref{eq:pc_integral} under the Gaussian assumption. It provides improved gradient signal in far-field regions while remaining faithful to collision risk ordering near closest approach. Similar Mahalanobis-based criteria have been used in robotics for uncertainty-aware collision avoidance \cite{akhtyamov2023mdmpc,zhang2021mdx}.

\subsection{Reward Design}

The per-step reward combines three components that penalize collision risk, propellant usage, and deviation from a reference trajectory:
\begin{equation}
r_t = r_{\text{risk}}(P_{c,t}; \tau_c) + r_{\text{fuel}}(\|\Delta\boldsymbol{v}\|_{1:t}; \tau_f) + \sum_{k=1}^{K} r_{\text{traj}}(|\Delta \theta_{t,k}|; \tau^{\text{traj}}_k),
\end{equation}
where $P_{c,t} \in [0,1]$ is the collision risk proxy, $\|\Delta\boldsymbol{v}\|_{1:t}$ is the cumulative velocity change, and $\Delta\boldsymbol{\theta}_t = [\Delta a_t, \Delta e_t, \Delta i_t, \Delta\Omega_t, \Delta\omega_t, \Delta M_t]$ are deviations of the spacecraft's osculating elements from their reference values. Angular differences are wrapped to $(-\pi, \pi]$.

To ensure that all reward components are dimensionless and comparable in magnitude, each scalar argument $x$ is mapped through a piecewise-linear shaping function with a component-specific threshold $\tau > 0$:
\begin{equation}
r(x; \tau) =
\begin{cases}
\displaystyle \frac{r_\tau}{\tau} x, & x \le \tau, \\[6pt]
\displaystyle r_\tau + \frac{r_{2\tau} - r_\tau}{2 - 1} \left(\frac{x}{\tau} - 1\right), & x > \tau,
\end{cases}
\qquad r_\tau = -1, \quad r_{2\tau} = -10.
\end{equation}
Under this shaping, reaching the threshold $\tau$ yields a penalty of $-1$, while reaching $2\tau$ yields $-10$. For $x > \tau$, the penalty continues to increase linearly with $x$; it reaches $-10$ at $x=2\tau$ and continues linearly beyond that point.

The three reward components are defined as follows:

\begin{itemize}
\item \emph{Collision risk.} Let $\boldsymbol{r}^{(S)}_t, \boldsymbol{v}^{(S)}_t$ and $\boldsymbol{r}^{(D)}_{t,j}, \boldsymbol{v}^{(D)}_{t,j}$ denote the inertial states of the spacecraft and debris $j$, respectively. The relative position is $\boldsymbol{\rho}_{t,j} = \boldsymbol{r}^{(D)}_{t,j} - \boldsymbol{r}^{(S)}_t$ and the relative velocity is $\boldsymbol{v}_{\text{rel},t,j} = \boldsymbol{v}^{(D)}_{t,j} - \boldsymbol{v}^{(S)}_t$. The environment identifies a conjunction set $\mathcal{D}_t$ containing all debris satisfying $\|\boldsymbol{\rho}_{t,j}\| \le R_{\text{crit}}$, with default $R_{\text{crit}} = 14$ km. Two formulations are available for the risk proxy, selectable at runtime:

\begin{itemize}
\item \emph{Mahalanobis kernel (default).} The risk kernel is computed in the encounter plane spanned by the transverse $T$ and out-of-plane $W$ axes as $k_{t,j} = \exp(-\frac{1}{2} \mathbf{u}_{t,j}^\top \Sigma_{TW,t}^{-1} \mathbf{u}_{t,j})$, where $\mathbf{u}_{t,j} = [\langle \boldsymbol{\rho}_{t,j}, \hat{\mathbf{T}}_{t,j} \rangle, \langle \boldsymbol{\rho}_{t,j}, \hat{\mathbf{W}}_{t,j} \rangle]$ and $\Sigma_{TW,t}$ is the $2 \times 2$ projection of the relative-position covariance $\Sigma_t$ from the UKF (specifically the position block $P_{xx}$). In addition to the conjunction set $\mathcal{D}_t$, a far-field set $\mathcal{F}_t = \{j : \|\boldsymbol{\rho}_{t,j}\| \le R_{\text{ff}}\} \setminus \mathcal{D}_t$ is included, with default $R_{\text{ff}} = 200$ km and at most $K_{\text{ff}} = 128$ nearest objects. The aggregate proxy is $P_{c,t} = 1 - \prod_{j \in \mathcal{D}_t \cup \mathcal{F}_t}(1 - k_{t,j})$.

\item \emph{Linearised Chen-Bai.} Per-debris collision probabilities $P_{c,t,j}$ are computed in closed form using the linearised two-dimensional formulation with joint covariance $\Sigma_{TW,t}$. These are aggregated via the union bound $1 - \prod_j(1 - P_{c,t,j})$, incorporating any prior probability mass accumulated earlier within the conjunction.
\end{itemize}

By default, the risk is evaluated on the observed states (after masking and noise) rather than the true states. The collision-risk contribution to the reward is $r_{\text{risk}}(P_{c,t}; \tau_c)$ with default threshold $\tau_c = 10^{-4}$.

\item \emph{Fuel usage.} The cumulative velocity change $\|\Delta\boldsymbol{v}\|_{1:t}$, computed from maneuver norms in m/s, enters the shaping function with threshold $\tau_f = 10$ m/s:
\begin{equation}
r_{\text{fuel}}(\|\Delta\boldsymbol{v}\|_{1:t}; \tau_f) = r(\|\Delta\boldsymbol{v}\|_{1:t}; \tau_f).
\end{equation}

\item \emph{Trajectory deviation.} Deviations in osculating elements are measured relative to a reference orbit, which defaults to the initial elements: $\Delta\boldsymbol{\theta}_t = \boldsymbol{\theta}_t - \boldsymbol{\theta}_{\text{ref}}$, with angle wrapping applied to $i$, $\Omega$, $\omega$, and $M$. Each element deviation is scaled by its own threshold and passed through the shaping function. The default thresholds are:
\begin{equation}
(\tau_a, \tau_e, \tau_i, \tau_\Omega, \tau_\omega, \tau_M) = (100~\text{m}, 10^{-2}, 10^{-2}~\text{rad}, 10^{-2}~\text{rad}, 10^{-2}~\text{rad}, \text{None}),
\end{equation}
where $\tau_M = \text{None}$ disables the mean anomaly penalty. This term encourages the policy to maintain mission-compatible orbits rather than allowing unconstrained trajectory perturbations.
\end{itemize}

At each step, the environment constructs the vector $$[P_{c,t}, \|\Delta\boldsymbol{v}\|_{1:t}, |\Delta a_t|, |\Delta e_t|, |\Delta i_t|, |\Delta\Omega_t|, |\Delta\omega_t|, |\Delta M_t|]$$ and applies the shaping function elementwise with the corresponding thresholds $$[\tau_c, \tau_f, \tau_a, \tau_e, \tau_i, \tau_\Omega, \tau_\omega, \tau_M]$$. The total reward is the sum of all shaped terms. If a runtime fault occurs, the episode terminates with an additional penalty of $-1$.

\subsection{Transformer-Based Policy Architecture}

Partially observable problems require policies that can extract information from sequences of observations rather than reacting to each observation in isolation. The choice of neural network architecture determines how historical observations are aggregated and what temporal dependencies can be captured.

Recurrent neural networks (RNNs) approach this by maintaining a hidden state $\mathbf{h}_t$ that is updated at each time step: $\mathbf{h}_t = \phi(W_x \mathbf{x}_t + W_h \mathbf{h}_{t-1} + \mathbf{b})$, where $\phi$ is a nonlinearity and $\mathbf{x}_t$ is the current input \cite{elman1990finding,rumelhart1986learning}. The hidden state is intended to summarize all relevant information from past observations. In practice, training RNNs via backpropagation through time leads to vanishing or exploding gradients, which limits their ability to capture long-range dependencies \cite{pascanu2013difficulty}. Long Short-Term Memory (LSTM) networks mitigate this through gating mechanisms that regulate information flow, allowing selective retention of relevant information over longer horizons \cite{hochreiter1997long}. However, both architectures compress the entire observation history into a fixed-size state vector, creating a representational bottleneck when the relevant past spans many time steps.

The Transformer architecture takes a different approach \cite{vaswani2017attention}. Rather than maintaining a compressed recurrent state, transformers retain representations of individual past observations and use self-attention to selectively weight their influence on the current output. Given an input sequence, each position computes queries, keys, and values through learned projections $Q = XW_Q$, $K = XW_K$, $V = XW_V$, and aggregates context as:
\begin{equation}
\mathrm{Attn}(Q, K, V) = \mathrm{softmax}\left(\frac{QK^\top}{\sqrt{d_k}}\right) V.
\end{equation}
Multi-head attention applies this computation in parallel with separate projections, and a position-wise feed-forward network processes the attended representations. Residual connections and layer normalization stabilize training in deep networks. Because each layer can attend over the full input sequence, information propagates across long ranges without passing through a bottleneck state. Positional encodings are added to preserve sequence order, as the attention mechanism is otherwise permutation-invariant.

Standard transformers are limited to a fixed context window determined at training time. Transformer-XL addresses this through segment-level recurrence \cite{Dai2019TXL}. The input is partitioned into segments; when processing segment $s$, the model caches the hidden activations from the preceding segment $s-1$ and makes them available as additional keys and values for attention in the current segment. Gradients are stopped through the cached memory to prevent backpropagation across the entire history. Relative positional encodings replace absolute position embeddings, allowing the attention computation to generalize when keys originate from earlier segments. This design extends the effective receptive field well beyond the segment length while retaining computational efficiency within each segment.

For collision avoidance under partial observability, the Transformer-XL architecture offers several advantages. When observations are intermittent or vary in quality, the attention mechanism can learn to weight informative observations more heavily regardless of when they occurred. The ability to selectively retrieve information from an extended history is particularly relevant when the agent must integrate observations gathered at different distances and under different sensing conditions. In our implementation, the Transformer-XL policy receives the observation sequence and outputs a Gaussian distribution over maneuver commands, with the mean specifying the commanded $\Delta\mathbf{v}_t$.

\section{Results}
\label{sec:results}

The experiments focus on single-debris encounters, chosen both as a computational simplification and as a controlled testbed for evaluating policy behavior under varying information quality. Encounter parameters are deliberately challenging: miss distances range from 0 to 50 metres, relative velocities span 1,000 to 10,000 m/s, and approach angles vary from 0 to 30 degrees. This distribution ensures that every episode requires an avoidance maneuver, exposing the agent to kinematically diverse scenarios.

Training proceeds with 64 parallel environments and rollouts of 256 steps, balancing exploration breadth against the depth needed for accurate credit assignment. Over a budget of $10^6$ environment steps with episode limits of 1,000 steps, the agent encounters at least 1,000 complete episodes while PPO performs approximately 60 policy updates. All experiments ran on a T4 GPU with expanded memory. The primary computational bottlenecks lie in the pykep library and Kalman filter computations, both of which are CPU-bound; more powerful GPUs offered marginal benefit.

Five observability regimes probe the effect of sensing degradation. These are defined over distance nodes at $[20,\, 100,\, 1{,}000,\, 100{,}000]$ km, with corresponding masking probabilities and noise levels specified in Table~\ref{tab:observability-regimes}. This unified framework spans conditions from near-perfect sensing to near-total information loss, enabling consistent comparisons across architectures and reward formulations.

\begin{table}[ht]
\centering
\caption{Observability regime definitions. Masking probability and noise standard deviation $\sigma$ are specified at four distance nodes: 20 km, 100 km, 1,000 km, and 100,000 km.}
\label{tab:observability-regimes}
\small
\begin{tabular}{lcccc}
\toprule
\textbf{Regime} & \multicolumn{4}{c}{\textbf{Mask probability / Noise $\sigma$}} \\
 & 20 km & 100 km & 1,000 km & 100,000 km \\
\midrule
Min & $10^{-6}$ / $10^{-6}$ & $10^{-5}$ / $10^{-5}$ & $10^{-4}$ / $10^{-4}$ & $10^{-3}$ / $10^{-3}$ \\
Low & $10^{-5}$ / $10^{-3}$ & $0.01$ / $0.01$ & $0.25$ / $0.05$ & $0.5$ / $0.1$ \\
Med & $10^{-4}$ / $10^{-3}$ & $0.4$ / $0.25$ & $0.75$ / $1.0$ & $0.9999$ / $2.0$ \\
High & $0.1$ / $0.01$ & $0.75$ / $0.75$ & $0.99$ / $4.0$ & $0.9999$ / $10.0$ \\
Max & $0.75$ / $0.74$ & $0.99$ / $4.0$ & $0.999$ / $10.0$ & $0.9999$ / $20.0$ \\
\bottomrule
\end{tabular}
\end{table}

Results are reported in two stages. The first stage analyses policies trained with a feed-forward multi-layer perceptron (MLP), a standard fully connected network that maps the current observation directly to an action without retaining memory of previous measurements. This provides a baseline for a decision policy that does not account for partial observability: under intermittent or noisy sensing, an MLP cannot aggregate information across time and must react to whatever information is available at the current step. The second stage reports results for the Transformer-XL (TrXL) policy, which augments decision-making with attention-based memory, and quantifies the benefits of memory by comparing TrXL against the MLP baseline.

\subsection{MLP Baseline: Reward Design and Observability Effects}
\label{sec:results-mlp}

\subsubsection*{Reward formulation comparison}

\begin{figure}[ht]
\centering
\begin{subfigure}[t]{0.48\textwidth}
    \centering
    \includegraphics[width=\textwidth]{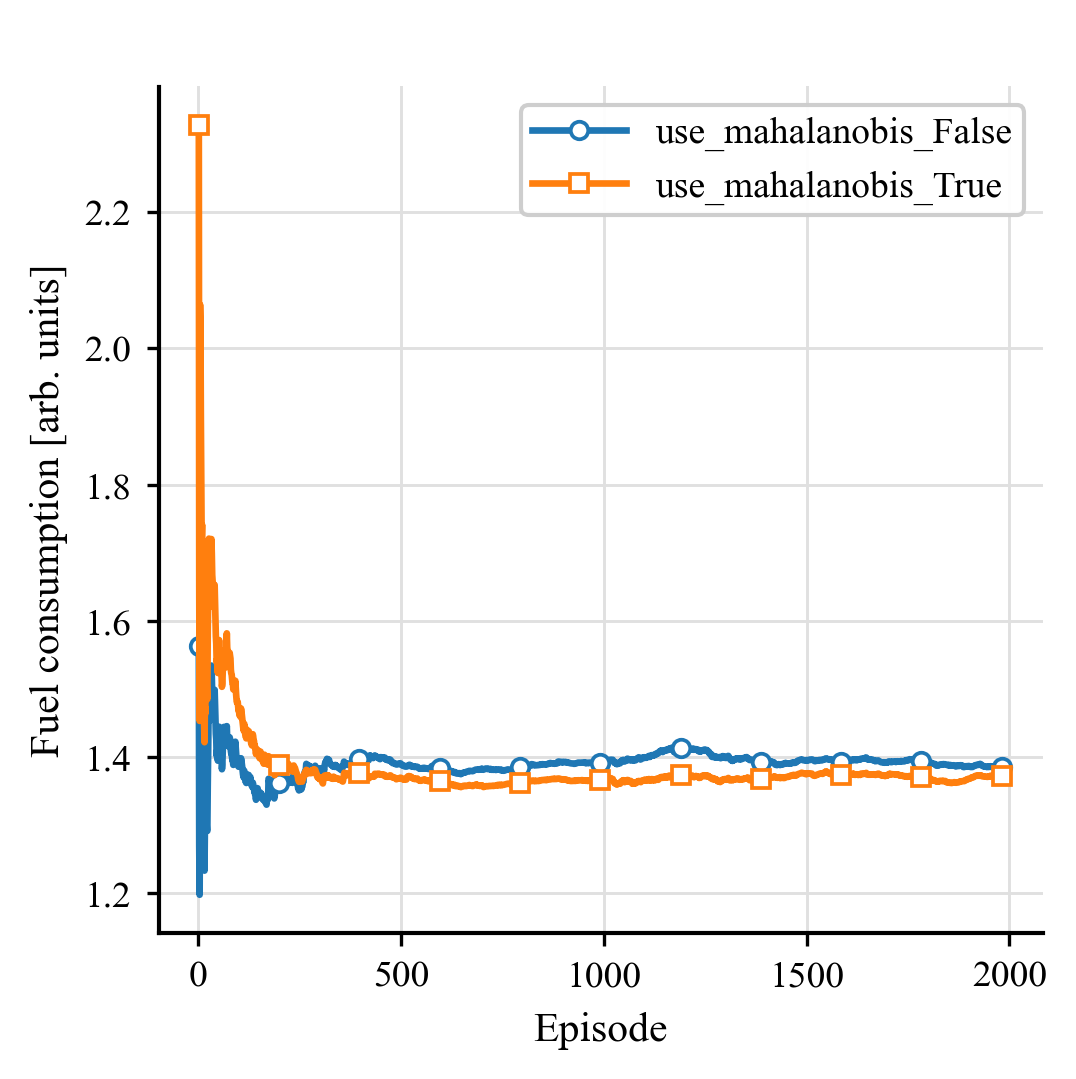}
    \caption{Fuel consumption.}
    \label{fig:fuel_compare_reward_type}
\end{subfigure}%
\hfill
\begin{subfigure}[t]{0.48\textwidth}
    \centering
    \includegraphics[width=\textwidth]{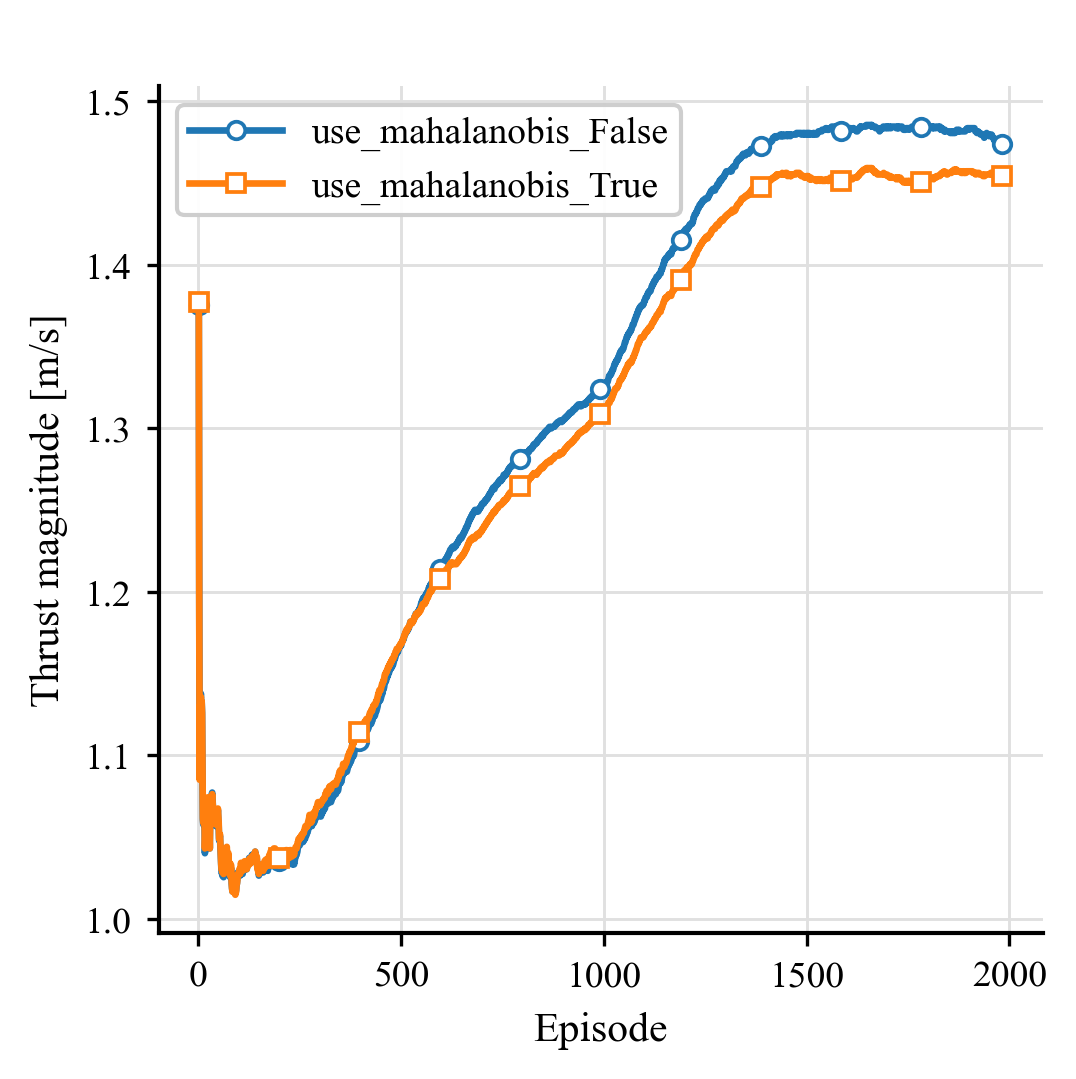}
    \caption{Thrust magnitude.}
    \label{fig:thrust_compare_reward_type}
\end{subfigure}
\caption{MLP training under medium observability: comparison of collision-risk reward formulations.}
\label{fig:compare_reward_type_both}
\end{figure}

The collision-risk term is a critical design choice. Two candidates are compared: the Chen-Bai linearized probability estimator and the Mahalanobis exponential kernel. This ablation is performed on the MLP baseline rather than on TrXL to isolate the effect of the reward formulation; introducing a memory mechanism would confound the comparison by simultaneously changing the policy class. Figure~\ref{fig:compare_reward_type_both} reports the evolution of fuel consumption during training, which serves as a sanity check that PPO is converging, and the impact of the reward choice on converged control effort.

Two observations follow from Figure~\ref{fig:compare_reward_type_both}. First, PPO training is stable for both rewards: in Figure~\ref{fig:fuel_compare_reward_type}, fuel cost drops from above 2.2 in the earliest episodes to a narrow band around 1.38--1.40 within roughly the first 200--300 episodes, then remains steady for the remainder of training. Second, the Mahalanobis reward yields consistently better efficiency at convergence. The converged fuel cost is approximately 1.38 with Mahalanobis versus 1.40 with Chen-Bai, a reduction of 1--2\%. A similar pattern appears in control effort: Figure~\ref{fig:thrust_compare_reward_type} shows the converged thrust magnitude is approximately 1.46 m/s for Mahalanobis compared to 1.48--1.49 m/s for Chen-Bai, roughly a 2\% reduction. In both cases, the agent learns collision-free behavior; the reward choice primarily affects the efficiency with which avoidance is achieved. These results, combined with the flexibility of the exponential kernel to tune risk sensitivity, motivate adopting the Mahalanobis formulation for all subsequent experiments.

\subsubsection*{Effect of partial observability}

\begin{figure}[ht]
\centering
\begin{subfigure}[t]{0.48\textwidth}
    \centering
    \includegraphics[width=\textwidth]{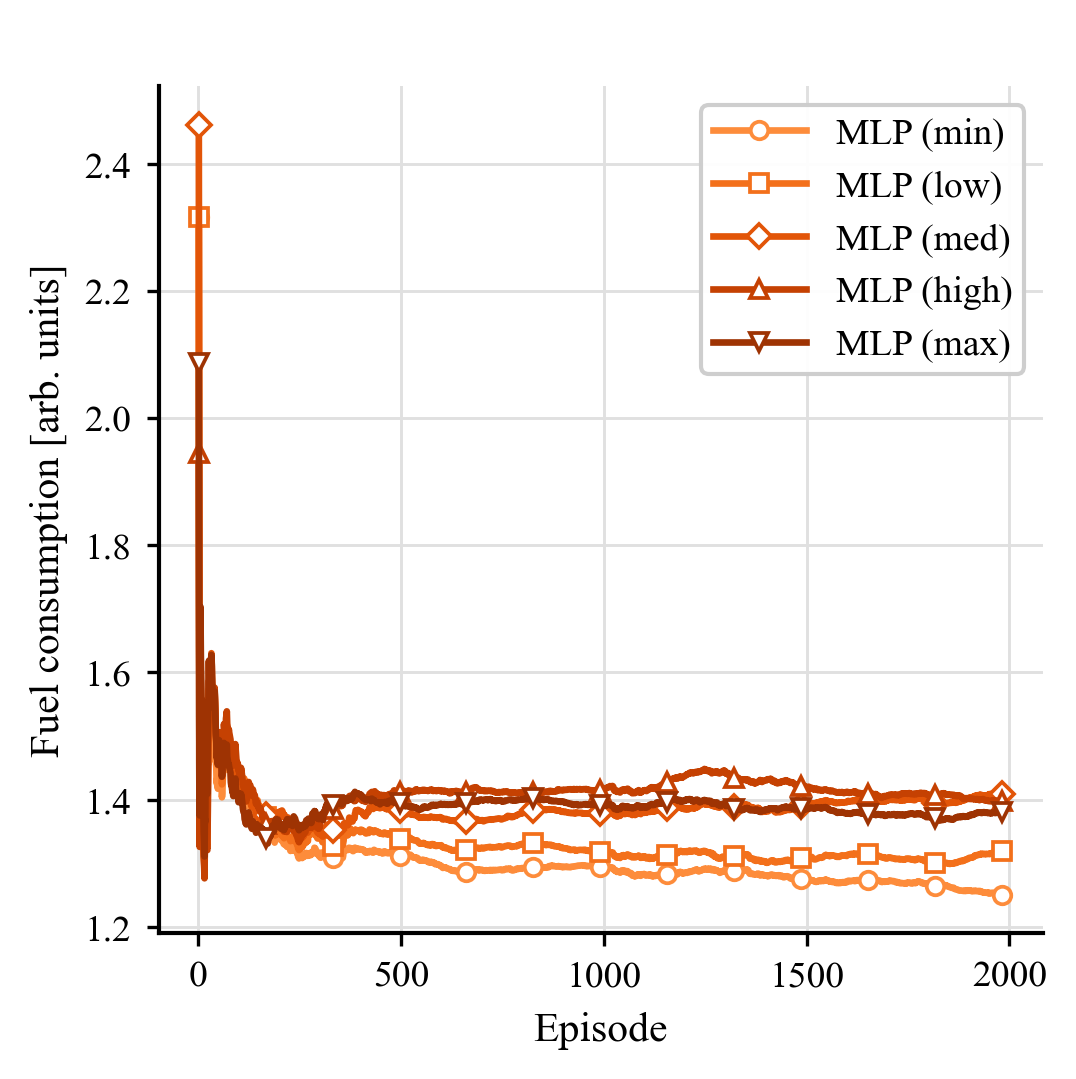}
    \caption{Fuel consumption.}
    \label{fig:compare_mode_partial}
\end{subfigure}%
\hfill
\begin{subfigure}[t]{0.48\textwidth}
    \centering
    \includegraphics[width=\textwidth]{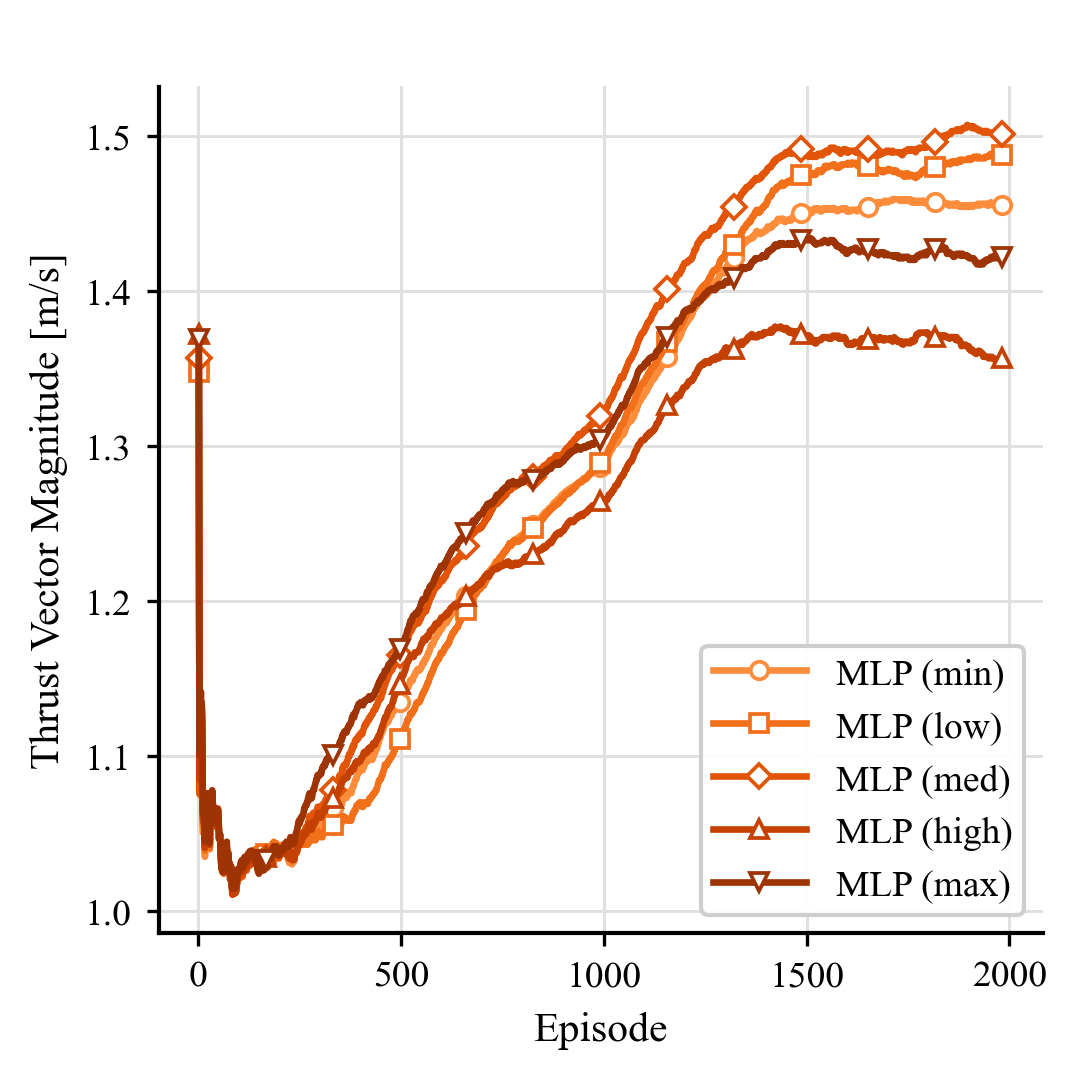}
    \caption{Thrust magnitude.}
    \label{fig:compare_core}
\end{subfigure}
\caption{MLP training dynamics across five partial-observability regimes.}
\label{fig:compare_both}
\end{figure}

With the reward formulation fixed, we examine how the MLP baseline responds to degraded sensing. Figure~\ref{fig:compare_both} shows fuel consumption and thrust magnitude during training across the five observability regimes.

Figure~\ref{fig:compare_mode_partial} reveals a clear separation of steady-state fuel cost across observability levels. Under minimal observability degradation (\texttt{min}), the MLP converges to a fuel cost of approximately 1.25. Under low degradation (\texttt{low}), it converges to approximately 1.32. Under more degraded sensing---medium (\texttt{med}), high (\texttt{high}), and maximal (\texttt{max})---fuel cost settles around 1.38--1.41. Concretely, \texttt{med} converges to approximately 1.41, representing a 13\% increase compared to \texttt{min} (Table~\ref{tab:train-end-results}, MLP rows). The thrust magnitude curves in Figure~\ref{fig:compare_core} converge smoothly across all regimes, ending in a band of approximately 1.35--1.50 m/s depending on the regime. These curves confirm that the MLP baseline degrades gracefully as observability decreases: it converges reliably but requires higher propellant expenditure under more severe masking and noise.

\begin{figure}[ht]
\centering
\begin{subfigure}[t]{0.48\textwidth}
    \centering
    \includegraphics[width=\textwidth]{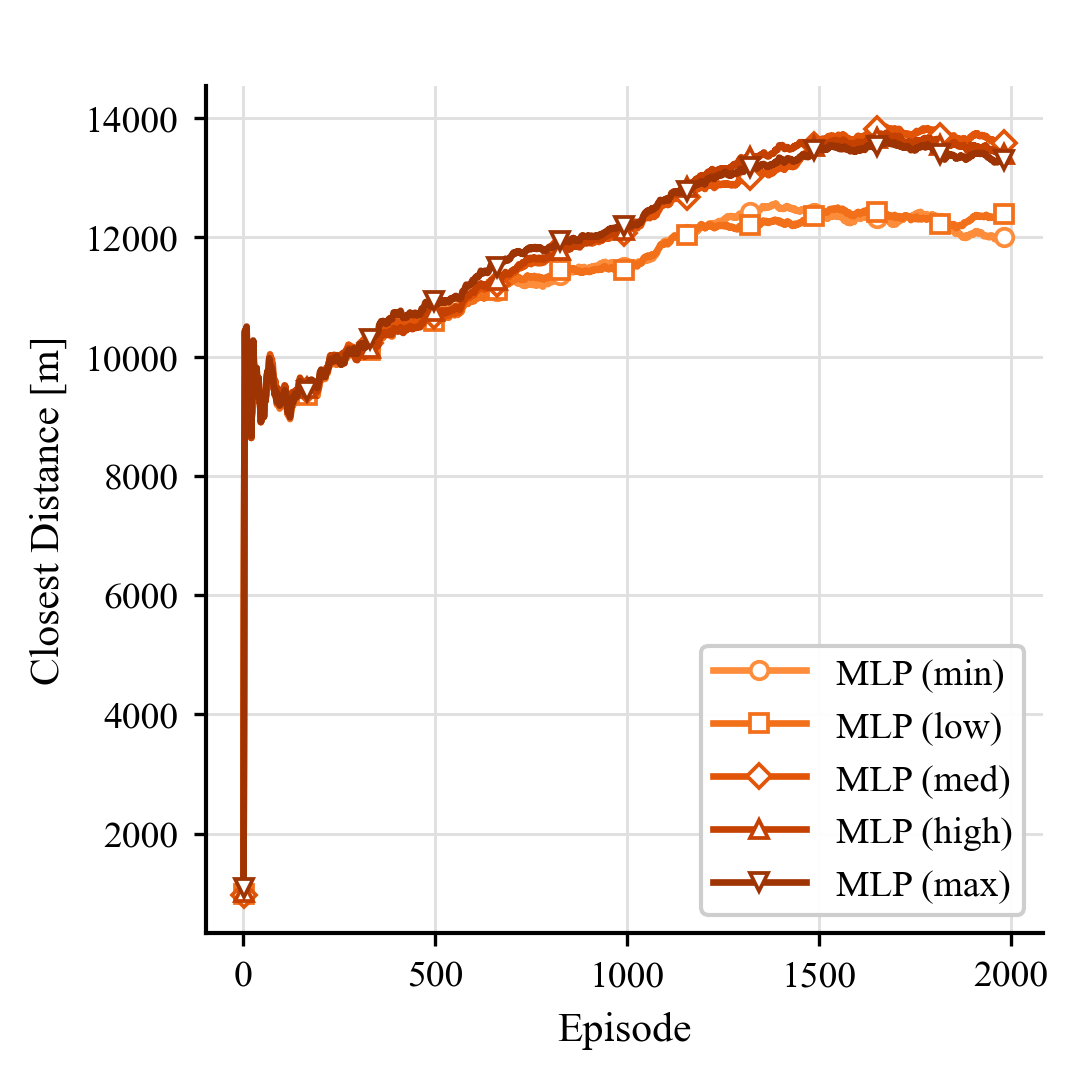}
    \caption{Closest approach distance.}
    \label{fig:mlp_dist_po}
\end{subfigure}%
\hfill
\begin{subfigure}[t]{0.48\textwidth}
    \centering
    \includegraphics[width=\textwidth]{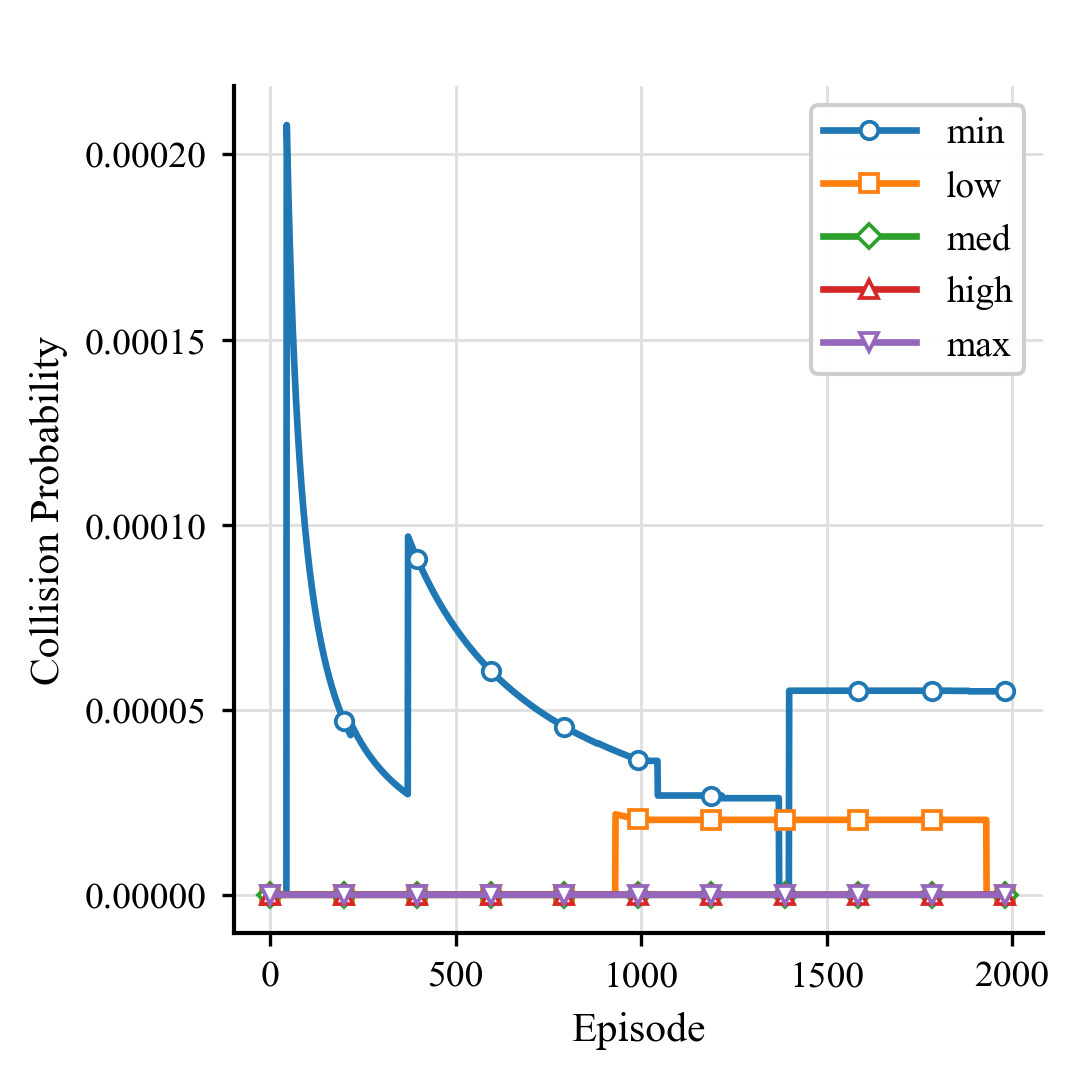}
    \caption{Estimated collision probability.}
    \label{fig:mlp_pc_po}
\end{subfigure}
\caption{MLP safety metrics across five partial-observability regimes.}
\label{fig:mlp_safety_po}
\end{figure}

Two additional metrics characterize safety performance: the closest approach distance achieved at conjunction and the estimated collision probability computed from the UKF mean and covariance. Figure~\ref{fig:mlp_safety_po} reports these for the MLP baseline across the five observability regimes.

Figure~\ref{fig:mlp_dist_po} shows that the converged closest approach increases from approximately 12.0 km under \texttt{min} to approximately 12.4 km under \texttt{low}, and reaches 13.3--13.6 km under \texttt{med}, \texttt{high}, and \texttt{max}. At \texttt{med}, the MLP converges to approximately 13.6 km, a 13\% larger stand-off distance than under \texttt{min} (Table~\ref{tab:train-end-results}).

Figure~\ref{fig:mlp_pc_po} reports the estimated collision probability. Across all regimes, the probability remains small and, by the end of training, stays below the $10^{-4}$ threshold used as a soft constraint in the reward. Under \texttt{min}, $P_c$ decreases from a transient peak of approximately $2.1 \times 10^{-4}$ to approximately $5.5 \times 10^{-5}$ at the end of training. Under \texttt{low}, the curve stabilizes around $2 \times 10^{-5}$. For \texttt{med}, \texttt{high}, and \texttt{max}, the estimated $P_c$ remains near the numerical floor (below $10^{-6}$), consistent with the larger closest-approach distances in Figure~\ref{fig:mlp_dist_po}.

Across all observability regimes, the collision probability constraint is satisfied in all episodes (Table~\ref{tab:train-end-results}), indicating that the MLP learns a robust avoidance policy even without memory. The MLP responds to degraded information by consuming more fuel and maintaining larger safety margins.

\subsection{Transformer-XL Performance Across Observability Regimes}
\label{sec:results-trxl-po}

We now analyse Transformer-XL (TrXL) policies trained under each observability regime. Unlike the MLP, TrXL maintains a fixed-length memory and uses attention to integrate information across time, which should help mitigate observation dropouts and time-varying noise.

\subsubsection*{Training dynamics}

\begin{figure}[ht]
\centering
\begin{subfigure}[t]{0.48\textwidth}
    \centering
    \includegraphics[width=\textwidth]{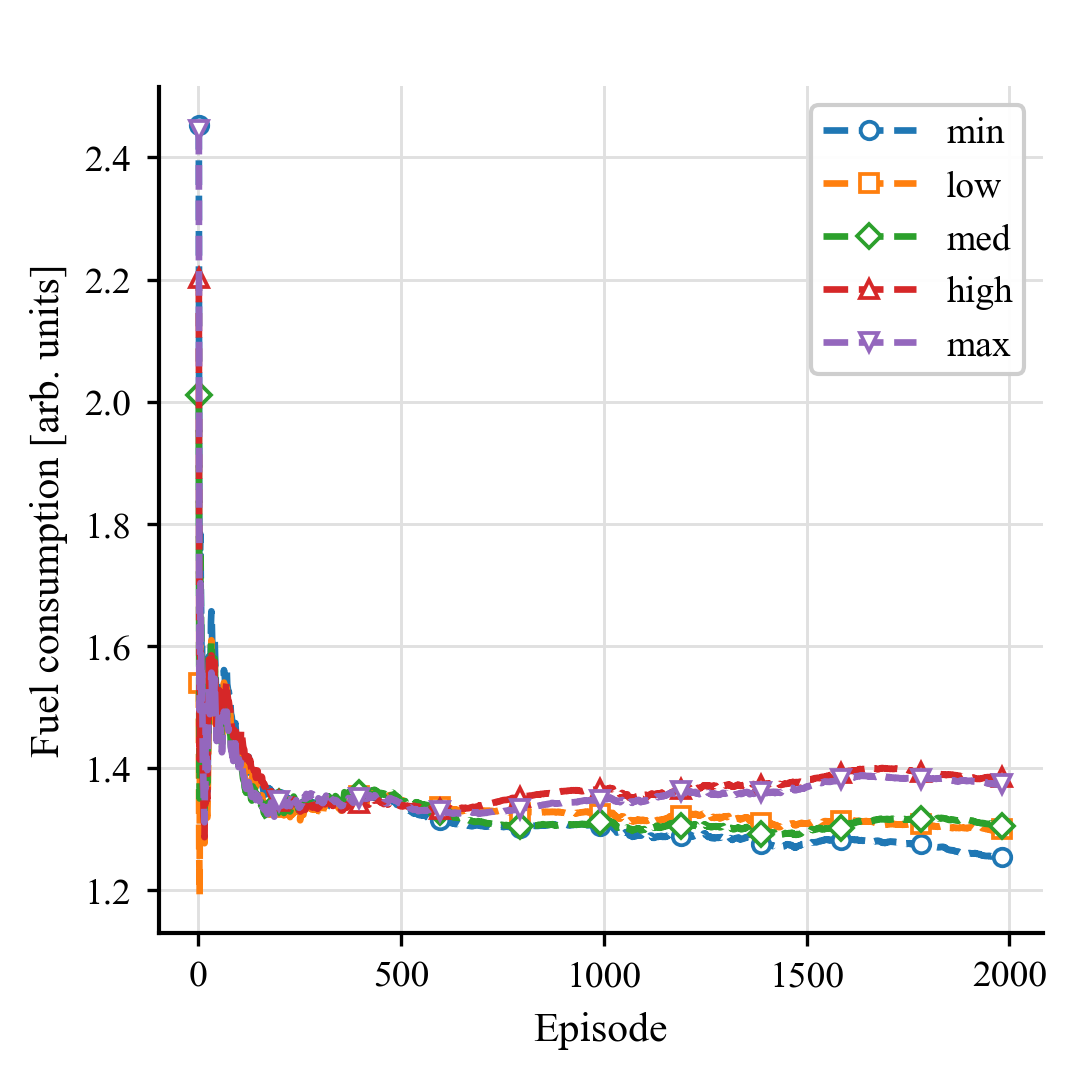}
    \caption{Fuel consumption.}
    \label{fig:trxl_fuel_po}
\end{subfigure}%
\hfill
\begin{subfigure}[t]{0.48\textwidth}
    \centering
    \includegraphics[width=\textwidth]{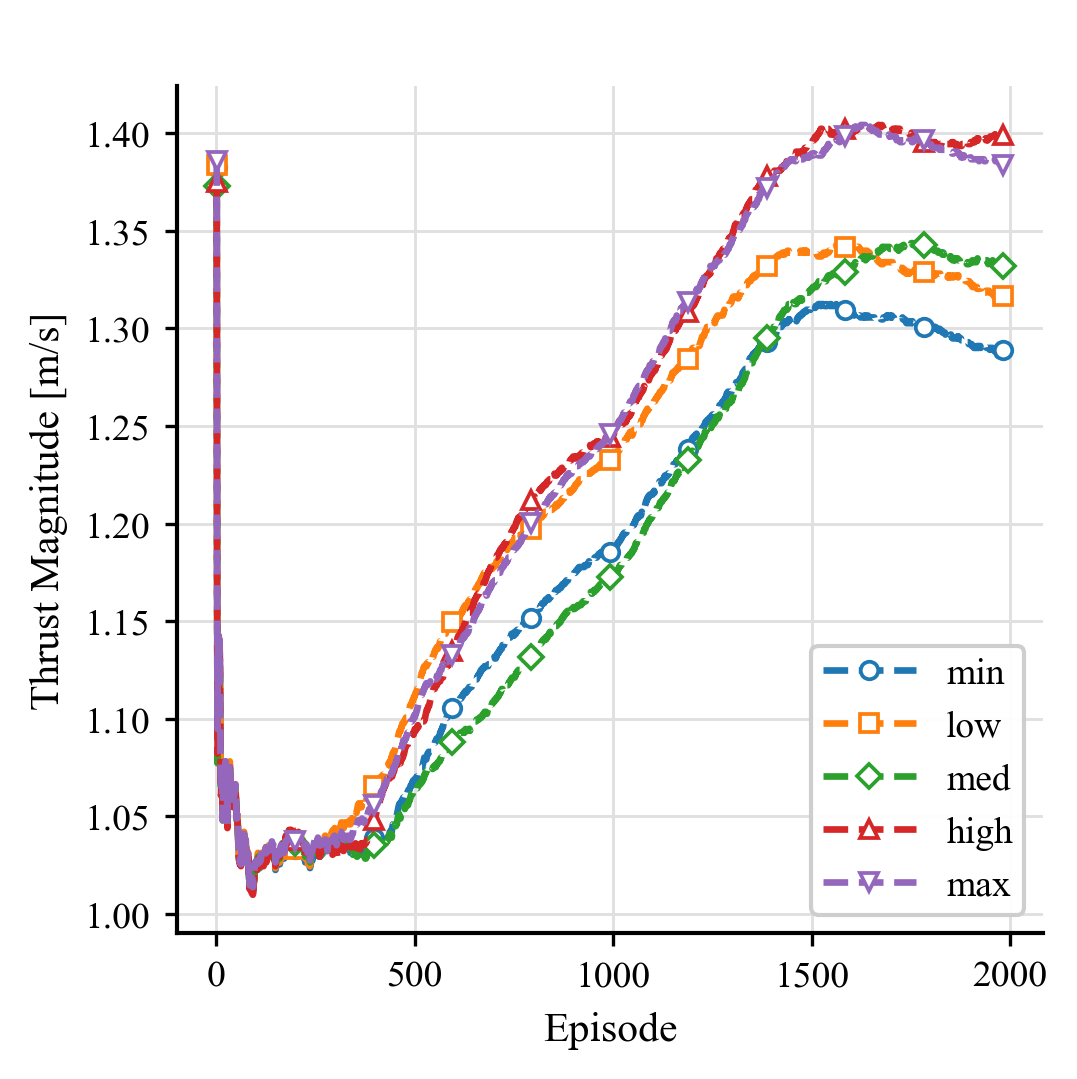}
    \caption{Thrust magnitude.}
    \label{fig:trxl_thrust_po}
\end{subfigure}
\caption{Transformer-XL training dynamics across five partial-observability regimes.}
\label{fig:trxl_train_po}
\end{figure}

Figure~\ref{fig:trxl_train_po} shows the evolution of fuel consumption and thrust magnitude during training for the five observability regimes.

After initial transients, fuel cost stabilizes in the 1.25--1.40 range depending on observability. Under \texttt{min}, TrXL converges to approximately 1.25, while \texttt{low} and \texttt{med} converge to approximately 1.30. Under the more degraded \texttt{high} and \texttt{max} settings, the converged fuel cost increases to approximately 1.38--1.40. The thrust magnitude curves exhibit the same ordering: \texttt{min} converges near 1.29 m/s, \texttt{low} and \texttt{med} near 1.32--1.34 m/s, and \texttt{high} and \texttt{max} near 1.39--1.40 m/s.

\subsubsection*{Safety metrics}

\begin{figure}[ht]
\centering
\begin{subfigure}[t]{0.48\textwidth}
    \centering
    \includegraphics[width=\textwidth]{
    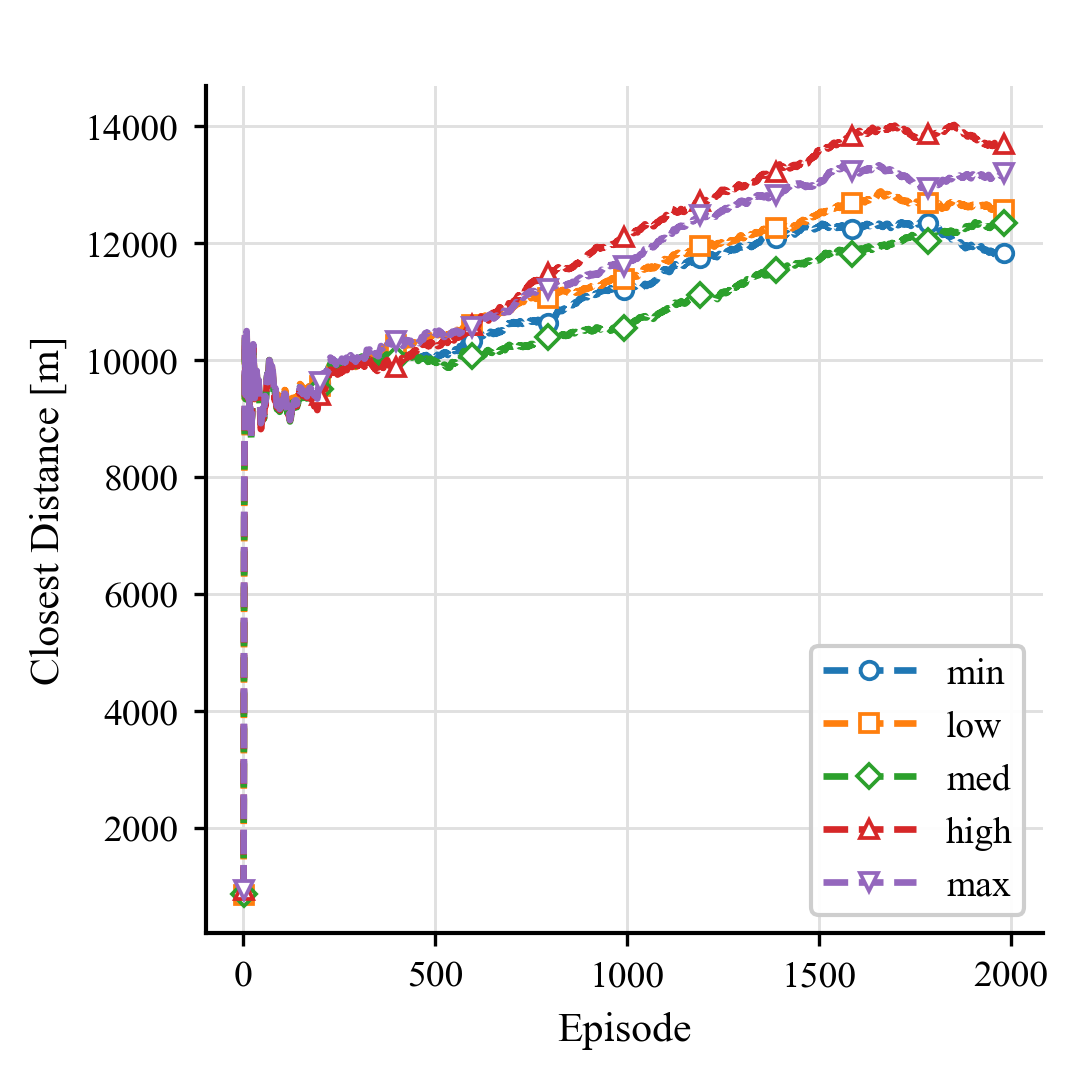}
    \caption{Closest approach distance.}
    \label{fig:trxl_dist_po}
\end{subfigure}%
\hfill
\begin{subfigure}[t]{0.48\textwidth}
    \centering
    \includegraphics[width=\textwidth]{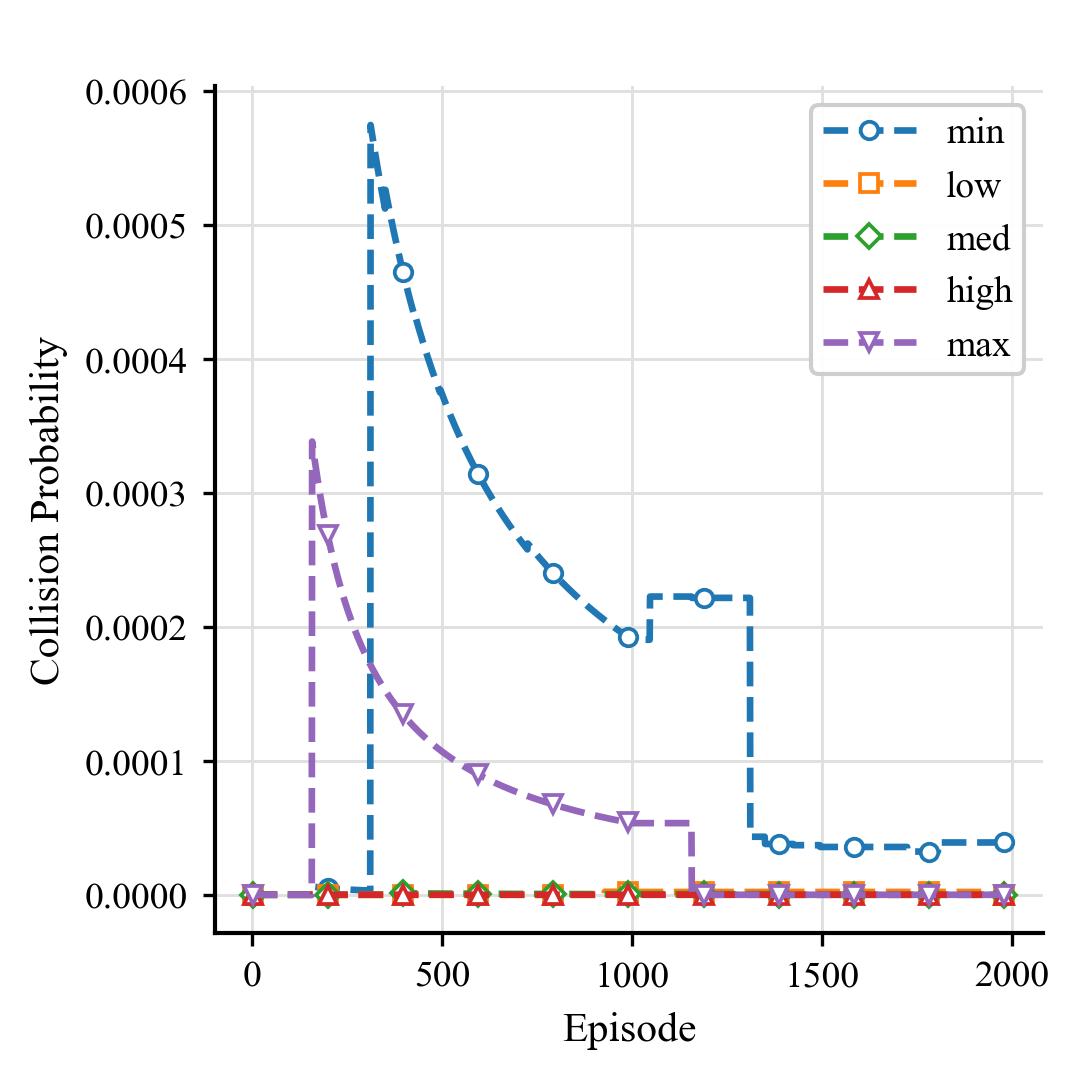}
    \caption{Estimated collision probability.}
    \label{fig:trxl_pc_po}
\end{subfigure}
\caption{Transformer-XL safety metrics across five partial-observability regimes.}
\label{fig:trxl_safety_po}
\end{figure}

Figure~\ref{fig:trxl_safety_po} reports the closest approach distance and the estimated collision probability across the five regimes.

In Figure~\ref{fig:trxl_dist_po}, the converged closest approach ranges from approximately 11.8 km under \texttt{min} to approximately 13.7 km under \texttt{high}, with \texttt{low} and \texttt{med} around 12.3--12.6 km and \texttt{max} around 13.2 km (Table~\ref{tab:train-end-results}). Figure~\ref{fig:trxl_pc_po} shows that the estimated collision probability remains small throughout training. Under \texttt{min}, $P_c$ decreases from approximately $5.8 \times 10^{-4}$ early in training to approximately $4 \times 10^{-5}$ by the end. Under \texttt{max}, the curve decreases from approximately $3.4 \times 10^{-4}$ to values below the plot resolution (below $10^{-6}$) after roughly 1,200 episodes. For \texttt{low}, \texttt{med}, and \texttt{high}, $P_c$ stays near the numerical floor throughout.

These results indicate that increasing partial observability shifts the TrXL policy toward more conservative behavior, with larger closest-approach distances and higher control effort. Training remains stable across all regimes, and the collision probability constraint is satisfied in all episodes.

\subsection{Architecture Comparison}
\label{sec:results-trxl}

We now compare Transformer-XL against the MLP baseline. The training curves focus on two representative observability levels (\texttt{med} and \texttt{high}) to maintain readability; the complete comparison across all five regimes is summarized in Table~\ref{tab:train-end-results}.

\subsubsection*{Training dynamics}

\begin{figure}[ht]
\centering
\begin{subfigure}[t]{0.48\textwidth}
    \centering
    \includegraphics[width=\textwidth]{
    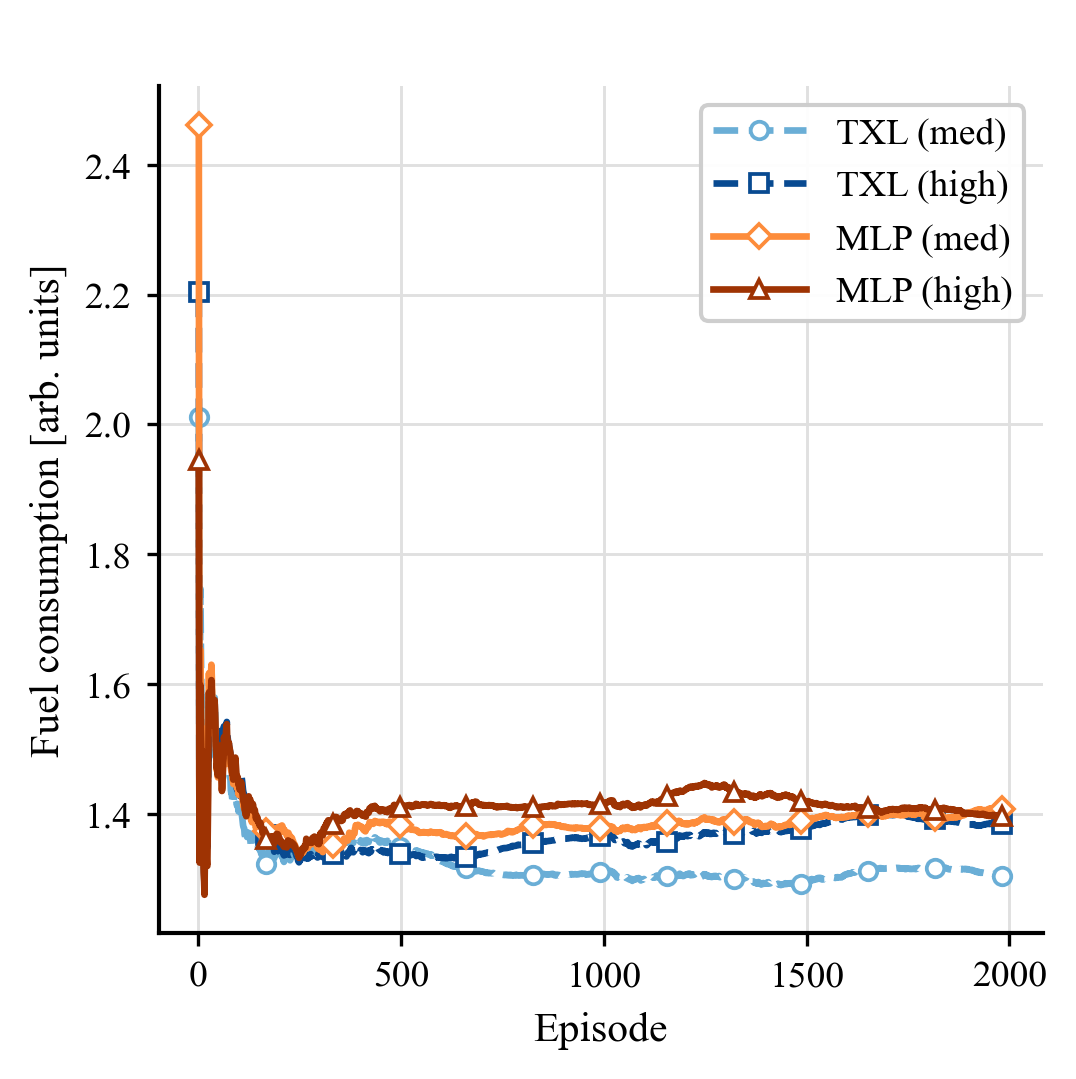}
    \caption{Fuel consumption (\texttt{med} and \texttt{high}).}
    \label{fig:arch_fuel}
\end{subfigure}%
\hfill
\begin{subfigure}[t]{0.48\textwidth}
    \centering
    \includegraphics[width=\textwidth]{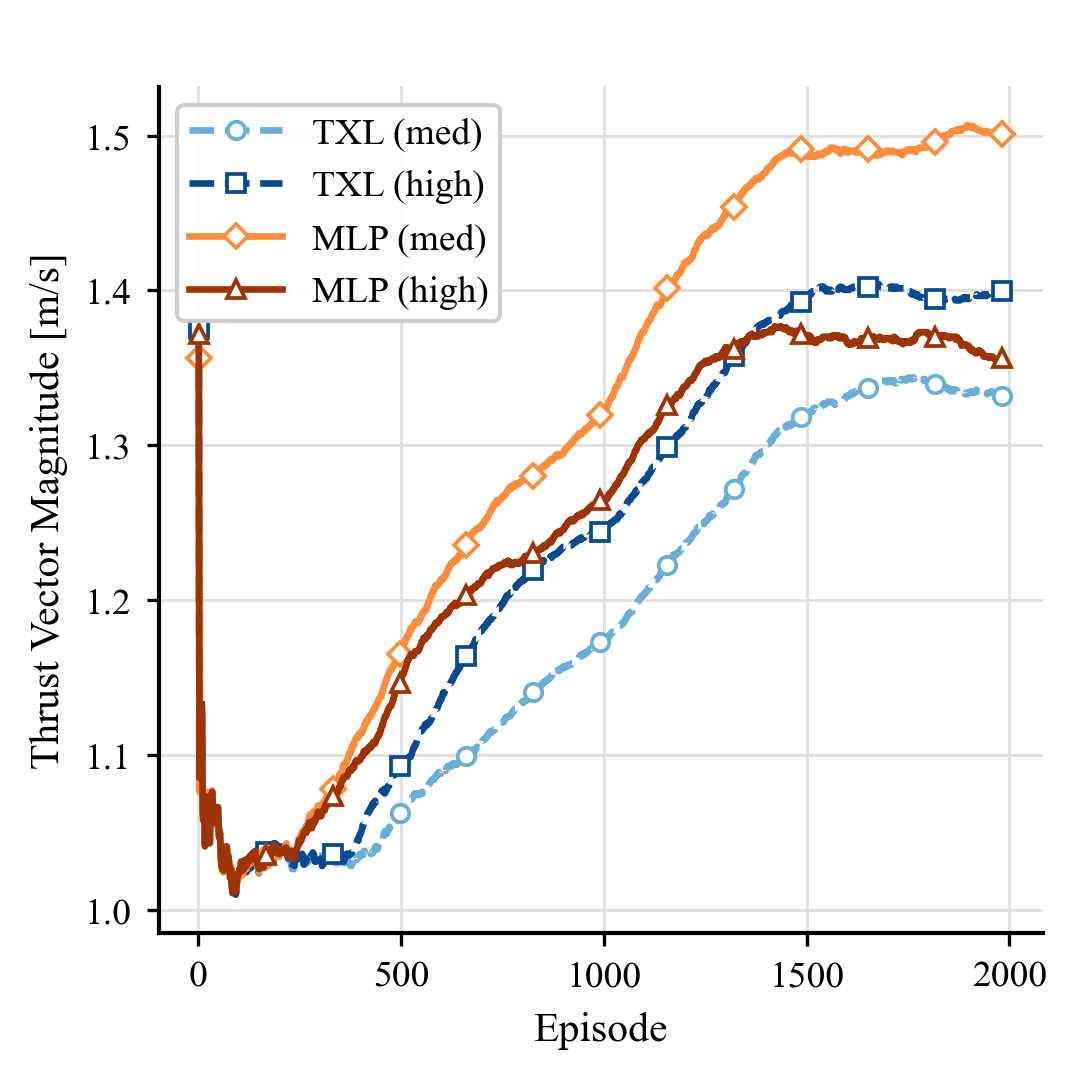}
    \caption{Thrust magnitude (\texttt{med} and \texttt{high}).}
    \label{fig:arch_thrust}
\end{subfigure}
\caption{MLP vs.\ Transformer-XL training dynamics under \texttt{med} and \texttt{high} observability.}
\label{fig:arch_comparison}
\end{figure}

Figure~\ref{fig:arch_comparison} compares fuel consumption and thrust magnitude for MLP and TrXL during training.

At \texttt{med} observability, TrXL converges to a fuel cost of approximately 1.30, whereas the MLP converges to approximately 1.40 (Figure~\ref{fig:arch_fuel}). Using the end-of-training averages from Table~\ref{tab:train-end-results}, this corresponds to a 7.8\% reduction in fuel cost (1.30 vs.\ 1.41). The same trend is evident in thrust magnitude (Figure~\ref{fig:arch_thrust}): TrXL stabilizes around 1.34 m/s, while the MLP stabilizes around 1.50 m/s, a reduction of approximately 10--11\%. At \texttt{high} observability, the fuel-cost gap narrows but remains in favor of memory: TrXL converges to approximately 1.38 versus 1.40 for the MLP, an improvement of 1--2\%. These results indicate that TrXL learns a more fuel-efficient avoidance strategy, with the advantage most pronounced in intermediate degradation regimes where the agent must routinely act with partial information.

\subsubsection*{Safety metrics}

\begin{figure}[ht]
\centering
\begin{subfigure}[t]{0.48\textwidth}
    \centering
    \includegraphics[width=\textwidth]{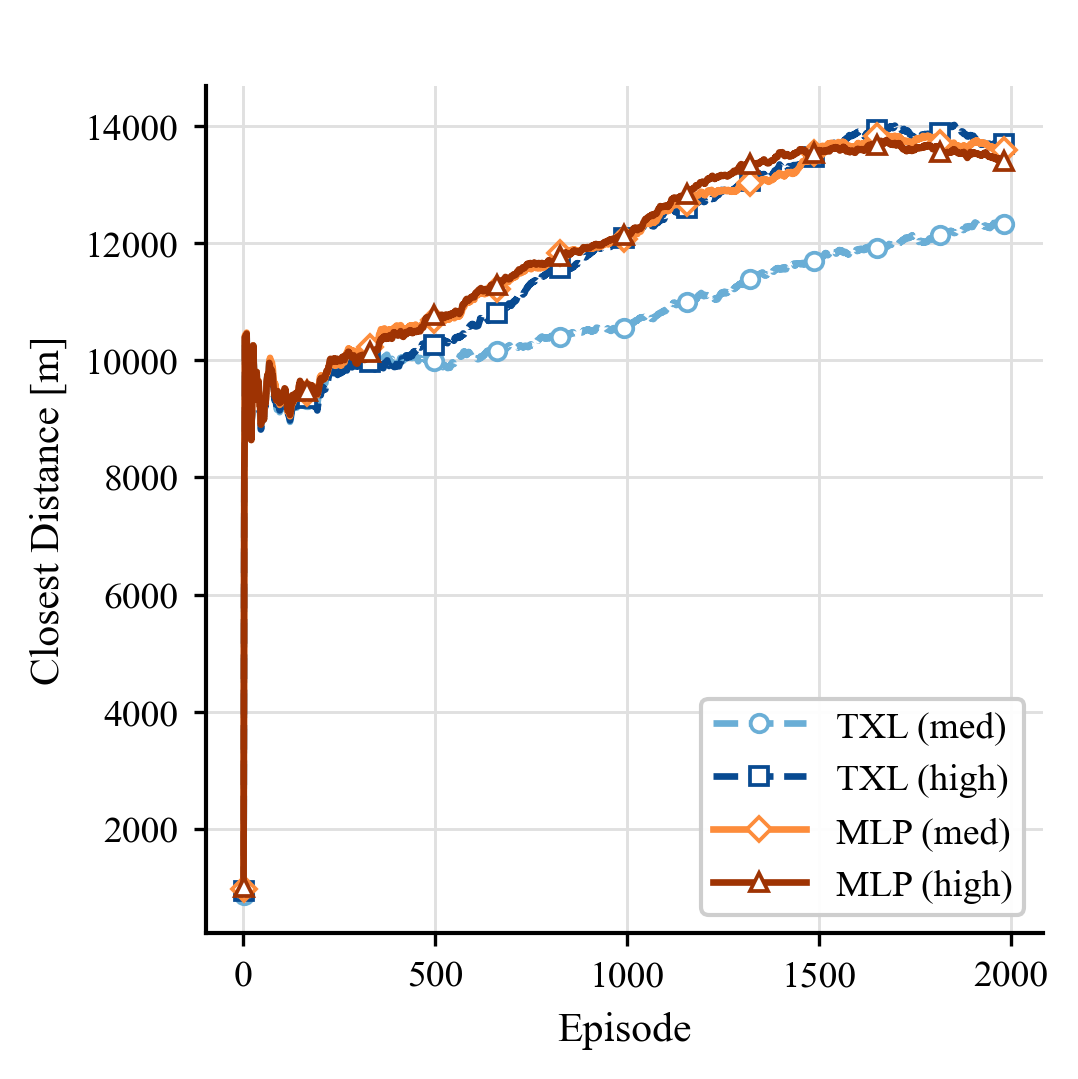}
    \caption{Closest approach distance (\texttt{med} and \texttt{high}).}
    \label{fig:dist_arch_po}
\end{subfigure}%
\hfill
\begin{subfigure}[t]{0.48\textwidth}
    \centering
    \includegraphics[width=\textwidth]{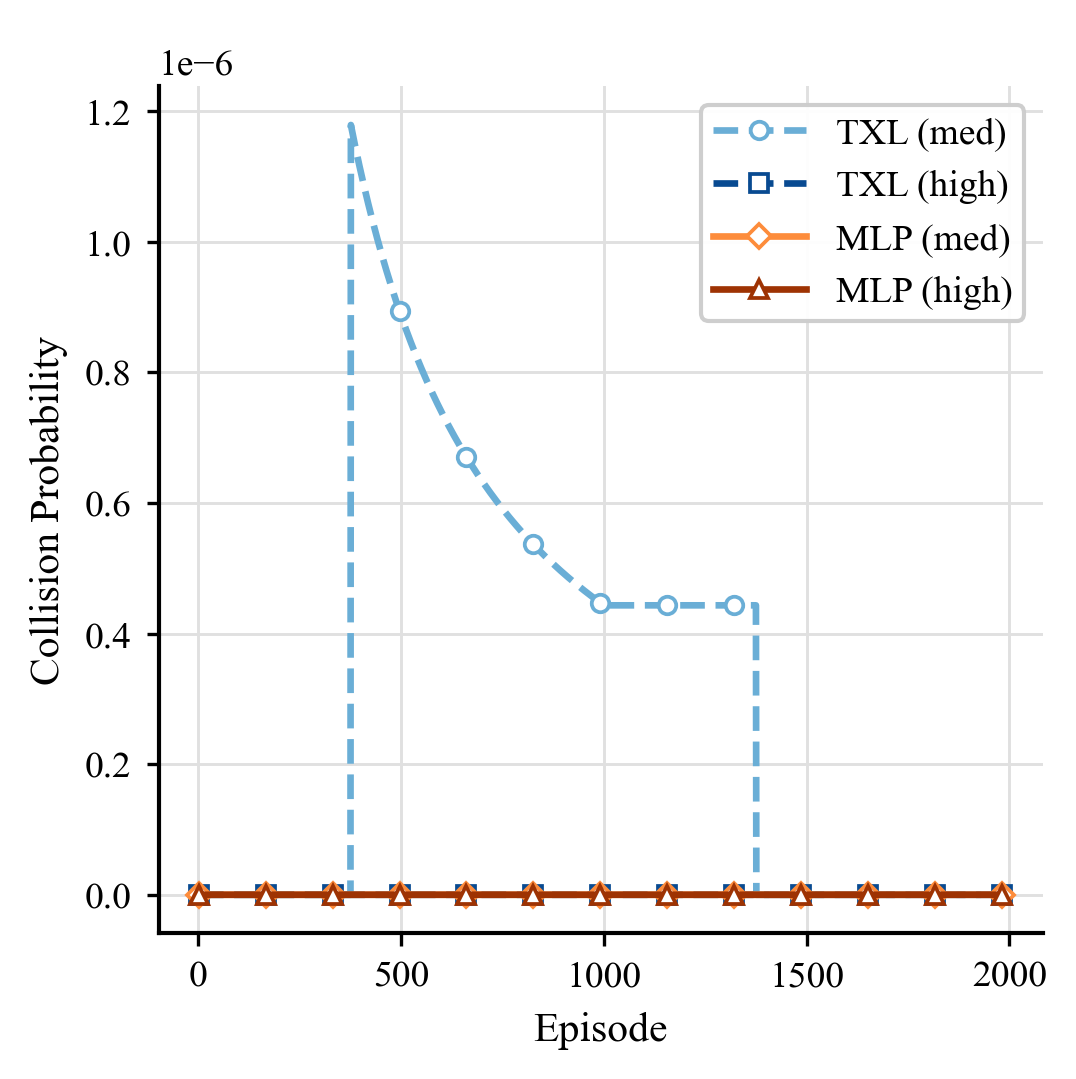}
    \caption{Estimated collision probability (\texttt{med} and \texttt{high}).}
    \label{fig:coll_prob_arch_po}
\end{subfigure}
\caption{MLP vs.\ Transformer-XL safety metrics under \texttt{med} and \texttt{high} observability.}
\label{fig:arch_safety_compare}
\end{figure}

Figure~\ref{fig:arch_safety_compare} compares closest approach distance and collision probability for the two architectures.

Figure~\ref{fig:dist_arch_po} compares closest distance during training. At \texttt{med}, TrXL converges to approximately 12.3 km while the MLP converges to approximately 13.6 km: TrXL maintains collision-free behavior while flying about 1.3 km closer, roughly a 10\% reduction in stand-off distance. At \texttt{high}, TrXL converges to approximately 13.7 km versus 13.4 km for the MLP, indicating a slightly more conservative margin in this harsher regime.

Figure~\ref{fig:coll_prob_arch_po} reports the corresponding collision probability estimates. Under both \texttt{med} and \texttt{high}, the values remain at or below $10^{-6}$, orders of magnitude below the $10^{-4}$ threshold used in the reward. The TrXL \texttt{med} curve exhibits a transient peak of approximately $1.1 \times 10^{-6}$ early in training and decreases below $5 \times 10^{-7}$ as training progresses; the remaining curves stay close to the numerical floor.

The pattern at \texttt{med} is particularly informative. TrXL achieves lower fuel consumption while maintaining a smaller closest approach distance, suggesting that memory enables more accurate risk assessment from intermittent observations. Rather than compensating for uncertainty through larger safety margins, the attention mechanism aggregates partial information over time to produce tighter, more confident maneuvers.

\subsubsection*{Summary across all regimes}

Table~\ref{tab:train-end-results} reports the converged performance for both architectures across all observability regimes, computed as rolling averages from the final training episodes. Both architectures satisfy the collision probability constraint ($P_c < 10^{-4}$) in all episodes across all regimes.

\begin{table}[ht]
\centering
\caption{End-of-training results (last-point rolling averages). Both architectures satisfy the collision probability constraint in all episodes.}
\label{tab:train-end-results}
\setlength{\tabcolsep}{6pt}
\renewcommand{\arraystretch}{1.15}
\begin{tabular}{llrrr}
\toprule
\textbf{Observability} & \textbf{Policy} & \textbf{Fuel Cost} & \textbf{Action (m/s)} & \textbf{Closest Distance (km)} \\
\midrule
Max & MLP & 1.38 & 2.45 & \textbf{13.3} \\
Max & TrXL & \textbf{1.37} & \textbf{2.42} & 13.2 \\
\midrule
High & MLP & 1.40 & \textbf{2.40} & 13.4 \\
High & TrXL & \textbf{1.38} & 2.43 & \textbf{13.7} \\
\midrule
Medium & MLP & 1.41 & 2.50 & \textbf{13.6} \\
Medium & TrXL & \textbf{1.30} & \textbf{2.39} & 12.3 \\
\midrule
Low & MLP & 1.32 & 2.47 & 12.4 \\
Low & TrXL & \textbf{1.30} & \textbf{2.38} & \textbf{12.6} \\
\midrule
Min & MLP & 1.25 & 2.44 & \textbf{12.0} \\
Min & TrXL & 1.25 & \textbf{2.37} & 11.8 \\
\bottomrule
\end{tabular}
\end{table}

Two conclusions emerge from these results. First, both architectures learn effective collision avoidance across all observability regimes. This outcome validates the framework design: the distance-dependent observation model provides a learnable structure for partial observability, the Mahalanobis reward yields stable gradients for policy optimization, and the UKF-based covariance estimation anchors the reward in operationally meaningful collision risk. Even a memoryless policy can exploit these components to achieve safe behavior.

Second, within this framework, Transformer-XL provides consistent efficiency gains. The improvement is largest at \texttt{med} (7.8\% fuel reduction) and smallest at the extremes: at \texttt{min}, near-perfect sensing leaves little information to aggregate over time; at \texttt{max}, observations are too sparse for even memory to compensate fully. The intermediate regimes---where information arrives intermittently but remains useful when aggregated---represent the operating conditions where attention-based temporal integration provides the greatest benefit. Across all regimes, TrXL matches or reduces action magnitudes (e.g., 2.50 to 2.39 m/s at \texttt{med}, a 4.4\% reduction), and closest distances remain in the operational band of 12--14 km.

The \texttt{med} regime illustrates the mechanism most clearly. TrXL achieves a tighter closest approach (12.3 km vs.\ 13.6 km) while consuming less fuel, indicating that memory allows the policy to extract more value from partial observations rather than defaulting to conservative behavior. This suggests that in realistic operational settings---where tracking data may be intermittent but not absent---transformer-based policies can reduce propellant expenditure without compromising safety.

\subsection{Out-of-Sample Evaluation}
\label{sec:results-oos}

To confirm that training-time advantages transfer to deployment, we evaluate each architecture--observability pair on 384 held-out episodes across 64 environments, matching the training distribution. Table~\ref{tab:sim-case-results} summarizes the results. Table~\ref{tab:sim-case-results} reports aggregate fuel usage, action magnitude, and closest-approach distances for each architecture and observability regime on this held-out set.

\begin{table}[ht]
\centering
\caption{Out-of-sample evaluation results. Both architectures satisfy the collision probability constraint in all episodes.}
\label{tab:sim-case-results}
\setlength{\tabcolsep}{6pt}
\renewcommand{\arraystretch}{1.15}
\begin{tabular}{llrrr}
\toprule
\textbf{Observability} & \textbf{Policy} & \textbf{Fuel Cost} & \textbf{Action (m/s)} & \textbf{Closest Distance (km)} \\
\midrule
Max & MLP & \textbf{1.41} & 1.49 & \textbf{13.3} \\
Max & TrXL & 1.44 & \textbf{1.40} & 12.1 \\
\midrule
High & MLP & 1.48 & 1.46 & 14.1 \\
High & TrXL & \textbf{1.44} & \textbf{1.38} & \textbf{14.8} \\
\midrule
Medium & MLP & 1.47 & 1.64 & \textbf{13.9} \\
Medium & TrXL & \textbf{1.38} & \textbf{1.38} & 12.8 \\
\midrule
Low & MLP & 1.32 & 1.57 & 11.7 \\
Low & TrXL & \textbf{1.30} & \textbf{1.39} & \textbf{12.3} \\
\midrule
Min & MLP & 1.32 & 1.49 & \textbf{11.5} \\
Min & TrXL & \textbf{1.27} & \textbf{1.34} & 11.0 \\
\bottomrule
\end{tabular}
\end{table}

The out-of-sample results corroborate the training-time findings. TrXL remains competitive or superior in efficiency across most regimes. At \texttt{med}, TrXL achieves a fuel cost of 1.38 versus 1.47 for the MLP (a 6.1\% reduction), and the mean action magnitude drops from 1.64 m/s to 1.38 m/s (a 15.9\% reduction). Similar patterns appear at \texttt{high} (fuel: 1.44 vs.\ 1.48, a 2.7\% reduction; action: 1.38 vs.\ 1.46 m/s, a 5.5\% reduction) and at \texttt{min} (fuel: 1.27 vs.\ 1.32, a 3.8\% reduction; action: 1.34 vs.\ 1.49 m/s, a 10.1\% reduction).

The only regime where TrXL incurs higher fuel cost is \texttt{max} (1.44 vs.\ 1.41, a 2.1\% increase). This reversal aligns with the interpretation from Section~\ref{sec:results-trxl}: under severe degradation, observations are too sparse for temporal aggregation to provide consistent benefit. Even so, TrXL maintains smaller action magnitudes (1.40 vs.\ 1.49 m/s) and satisfies the collision constraint in all episodes.

Two aspects of the out-of-sample results merit emphasis. First, the efficiency gains observed during training generalize to held-out encounters, indicating that the learned policies capture transferable structure rather than overfitting to specific trajectories. Second, the action magnitudes for TrXL are systematically lower across nearly all regimes, suggesting smoother control profiles that could reduce actuator wear in operational deployment. The consistent satisfaction of the collision probability constraint across both architectures and all observability conditions confirms that the framework produces robust avoidance behavior that does not degrade when evaluated on new encounter geometries.

\section{Conclusion}
\label{sec:conclusion}

This work develops a reinforcement learning framework for autonomous collision avoidance that addresses the information challenges inherent in orbital operations. The framework integrates four components typically treated separately in conjunction analysis into a closed-loop simulation with a consistent reinforcement learning interface, namely (i) orbital propagation, (ii) state estimation, (iii) collision risk evaluation, and (iv) decision-making,

The primary methodological contribution is the formulation of collision avoidance as a learning problem under distance-dependent partial observability. Detection probability and measurement quality are modeled as explicit functions of range through Lagrange interpolation, capturing the operational reality that tracking reliability varies with geometry. This formulation, combined with a Mahalanobis-based reward surrogate anchored in UKF covariance estimates, enables stable policy optimization while maintaining operational interpretability. The experimental results validate this design showing that both feedforward and memory-augmented architectures learn effective avoidance across observability regimes spanning near-perfect tracking to severe information loss. The fact that even a memoryless policy achieves collision-free behavior in all tested scenarios confirms that the framework provides learnable structure for the partial observability problem.

Within this framework, Transformer-XL policies provide consistent efficiency gains over the MLP baseline. The improvement concentrates in intermediate degradation regimes, where observations arrive intermittently but remain useful when aggregated over time. At medium observability, TrXL reduces fuel cost by 7.8\% while maintaining a tighter closest approach, indicating that attention-based temporal integration enables more accurate risk assessment rather than compensating for uncertainty through conservative margins. The gains diminish at extremes where near-perfect sensing leaves little to aggregate, while severe degradation provides too few observations for memory to exploit. Out-of-sample evaluation confirms that these advantages generalize to held-out encounters.

Several limitations qualify these findings. The experiments consider single-debris encounters; multi-object scenarios would introduce combinatorial complexity in threat prioritization and maneuver sequencing that may alter the relative benefits of memory. The two-body Keplerian dynamics neglect perturbations ($J_2$, atmospheric drag, solar radiation pressure) that affect covariance growth over longer time horizons. The collision probability constraint is enforced through reward shaping rather than hard constraints, providing empirical but not formal safety guarantees. Finally, the observability model assumes known degradation profiles; in practice, the mapping from range to observation quality may itself be uncertain or time-varying.

These limitations suggest directions for further work. Extending the framework to multi-debris encounters would test whether learned policies can handle competing threats and downstream constraints on maneuver planning and higher-fidelity dynamics would improve realism for operational deployment.

\newpage
\section*{Appendix}

\makeatletter
\let\cleardoublepage\clearpage
\let\cleardoubleoddpage\clearpage
\makeatother
\appendix

\subsection*{Notation}
\setlength{\tabcolsep}{4pt}
\renewcommand{\arraystretch}{1.0}

\begin{tabularx}{\linewidth}{@{}l l X@{}}
\toprule
\textbf{Symbol} & \textbf{Units} & \textbf{Definition} \\
\midrule
\multicolumn{3}{@{}l}{\textit{Orbital mechanics}} \\[2pt]
$\bm{r}^{(S)},\,\bm{r}^{(D)}$ & km & Inertial position of servicer and debris \\
$\delta\bm{r}_t,\,\delta\bm{v}_t$ & km, km/s & Relative position and velocity in LVLH frame \\
$\bm{x}_t = [\delta\bm{r}_t;\,\delta\bm{v}_t]$ & -- & Relative state vector \\
$\Delta\bm{v}_t$ & m/s & Impulsive maneuver (action) \\
$d_{\mathrm{miss}}$ & km & Miss distance at closest approach \\
\midrule
\multicolumn{3}{@{}l}{\textit{Reinforcement learning}} \\[2pt]
$\bm{o}_t$ & -- & Observation (masked and noisy state) \\
$r_t$ & -- & Scalar reward at time $t$ \\
$\gamma$ & -- & Discount factor \\
$\pi_\theta(\bm{a}_t \mid \bm{o}_t)$ & -- & Policy parameterized by $\theta$ \\
$V^\pi,\,A^\pi$ & -- & Value and advantage functions \\
\midrule
\multicolumn{3}{@{}l}{\textit{State estimation and collision risk}} \\[2pt]
$\hat{\bm{x}}_{t|t}$ & -- & Kalman filter state estimate \\
$\bm{P}_{t|t}$ & -- & State covariance matrix \\
$\bm{\Sigma}_{TW}$ & km$^2$ & Encounter-plane covariance \\
$d_M^2$ & -- & Squared Mahalanobis distance \\
$P_c$ & -- & Collision probability \\
\bottomrule
\end{tabularx}

\vspace{1ex}
\noindent\textit{Conventions:} Positions in km, velocities in km/s (m/s for $\Delta v$). LVLH denotes the Local Vertical, Local Horizontal frame with radial ($R$), along-track ($T$), and cross-track ($W$) axes.

\newpage
\begin{figure}[h]
\centering
\caption{Summary of the framework being used}
\includegraphics[width=0.85\textwidth]{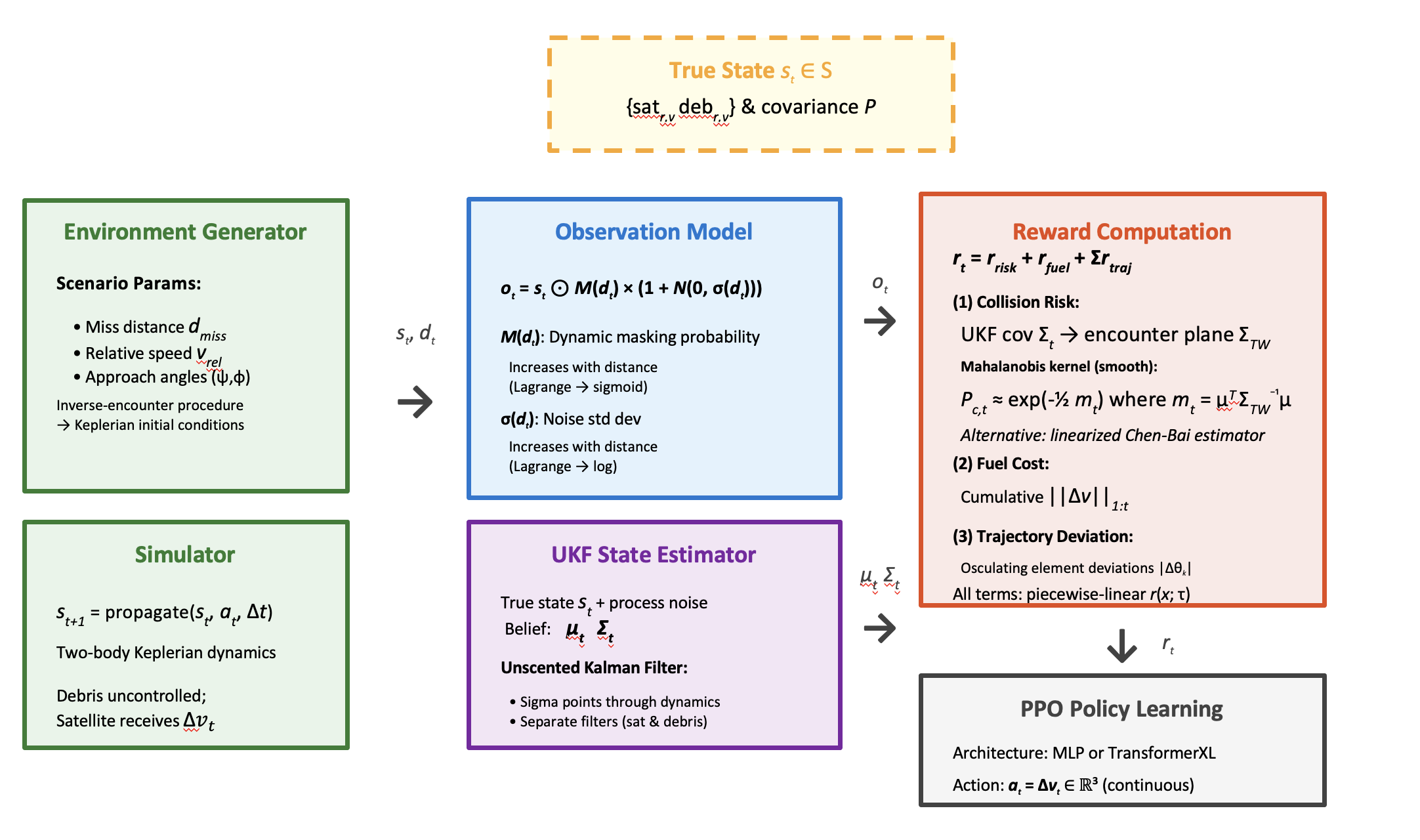}
\end{figure}

\begin{table}[ht]
\centering
\caption{PPO hyperparameters.}
\label{tab:ppo-hparams}
\footnotesize
\setlength{\tabcolsep}{4pt}
\renewcommand{\arraystretch}{0.95}
\begin{tabular}{lr@{\hspace{1.5em}}lr}
\toprule
\textbf{Parameters} & \textbf{Value} & \textbf{Parameters} & \textbf{Value} \\
\midrule
Discount factor $\gamma$     & 0.995          & Max grad norm             & 0.25 \\
GAE $\lambda$                & 0.95           & Actor learning rate (init) & $2.75\times10^{-4}$ \\
Update epochs                & 3              & Critic learning rate (init) & $2.75\times10^{-4}$ \\
Num. environments            & 64             & Final learning rate       & $1.0\times10^{-5}$ \\
Rollout length (steps / env) & 256            & LR/entropy anneal steps   & $1.5\times10^{6}$ \\
Rollout batch size           & 16{,}384       & Total timesteps           & $1.0\times10^{6}$ \\
Num. minibatches             & 8              & Target KL                 & --- \\
Minibatch size\textsuperscript{$\dagger$} & 2{,}048 & Value function clipping & True \\
Clipping ratio               & 0.25           & Normalize advantages      & True \\
Entropy coeff. (init)        & $1.0\times10^{-3}$ & Action squashing       & True \\
Entropy coeff. (final)       & $5.0\times10^{-4}$ &                          &      \\
Value loss coeff.            & 0.5            &                          &      \\
\bottomrule
\end{tabular}

\raggedright\footnotesize
\textsuperscript{$\dagger$}\,Derived as \(\text{num\_envs}\times \text{num\_steps} \,/\, \text{num\_minibatches} = 64\times256/8 = 2048\).
\end{table}

\section*{Acknowledgments}
This work was supported by the \emph{Agence Nationale de la Recherche (ANR)}, France, under the project \textbf{ANR-23-CE10-0006} (Resilient and Sustainable Planning and Management of Future Space Industry Infrastructure with On-Orbit Servicing) for which Dr. Adam Abdin is the principal investigator.

\bibliographystyle{elsarticle-num-names}
\bibliography{refs.bib}

\end{document}